\documentclass{article}

\usepackage{microtype}
\usepackage{graphicx}
\usepackage{subfigure}
\usepackage{booktabs} 
\usepackage{multirow}
\usepackage[utf8]{inputenc}
\usepackage{pifont}

\usepackage{cancel}

\usepackage{hyperref}

\usepackage[accepted]{icml2025}

\usepackage{algorithm}
\usepackage{amsmath}
\usepackage{amssymb}
\usepackage{mathtools}
\usepackage{amsthm}

\usepackage{xcolor}
\usepackage{listings}
\usepackage{float}

\usepackage[capitalize,noabbrev]{cleveref}

\theoremstyle{plain}

\theoremstyle{definition}

\theoremstyle{remark}

\usepackage{subcaption}

\usepackage[textsize=tiny]{todonotes}

\icmltitlerunning{InfoSAM: Fine-Tuning the Segment Anything Model from An Information-Theoretic Perspective}

\begin{document}

\twocolumn[
\icmltitle{InfoSAM: Fine-Tuning the Segment Anything Model \\ from An Information-Theoretic Perspective}



\icmlsetsymbol{equal}{*}

\begin{icmlauthorlist}
\icmlauthor{Yuanhong Zhang}{equal,sch,MOEKLINNS}
\icmlauthor{Muyao Yuan}{equal,sch,MOEKLINNS}
\icmlauthor{Weizhan Zhang\textsuperscript{\textdagger}}{sch,MOEKLINNS}
\icmlauthor{Tieliang Gong}{sch,BDKE}
\icmlauthor{Wen Wen}{sch,BDKE}
\icmlauthor{Jiangyong Ying}{Tele}
\icmlauthor{Weijie Shi}{sch2}
\end{icmlauthorlist}

\icmlaffiliation{sch}{School of Computer Science and Technology, Xi’an Jiaotong University, Xi’an, China}
\icmlaffiliation{MOEKLINNS}{Ministry of Education Key Laboratory of Intelligent Networks and Network Security, Xi’an Jiaotong University, Xi’an, China}
\icmlaffiliation{BDKE}{Shaanxi Province Key Laboratory of Big Data Knowledge Engineering, Xi’an Jiaotong University, Xi’an, China}
\icmlaffiliation{Tele}{China Telecom E-surfing Vision Technology Co., Ltd, Hangzhou, China}
\icmlaffiliation{sch2}{School of Electrical Engineering, Xi’an Jiaotong University, Xi’an, China}

\icmlcorrespondingauthor{\textsuperscript{\textdagger}Weizhan Zhang}{zhangwzh@xjtu.edu.cn}

\icmlkeywords{Machine Learning, ICML}

\vskip 0.3in
]



\printAffiliationsAndNotice{\icmlEqualContribution} 

\begin{abstract}
The Segment Anything Model (SAM), a vision foundation model, exhibits impressive zero-shot capabilities in general tasks but struggles in specialized domains.
Parameter-efficient fine-tuning (PEFT) is a promising approach to unleash the potential of SAM in novel scenarios. 
However, existing PEFT methods for SAM neglect the domain-invariant relations encoded in the pre-trained model. 
To bridge this gap, we propose InfoSAM, an information-theoretic approach that enhances SAM fine-tuning by distilling and preserving its pre-trained segmentation knowledge.
Specifically, we formulate the knowledge transfer process as two novel mutual information-based objectives: (i) to compress the domain-invariant relation extracted from pre-trained SAM, excluding pseudo-invariant information as possible, and (ii) to maximize mutual information between the relational knowledge learned by the teacher (pre-trained SAM) and the student (fine-tuned model). 
The proposed InfoSAM establishes a robust distillation framework for PEFT of SAM.
Extensive experiments across diverse benchmarks validate InfoSAM's effectiveness in improving SAM family's performance on real-world tasks, demonstrating its adaptability and superiority in handling specialized scenarios.
The code and models are available at \href{https://muyaoyuan.github.io/InfoSAM_Page}{InfoSAM project page}.

\end{abstract}

\section{Introduction}
\label{introduction}

\begin{figure}[t]
\vskip 0.2in
\begin{center}
\centerline{\includegraphics[width=\columnwidth]{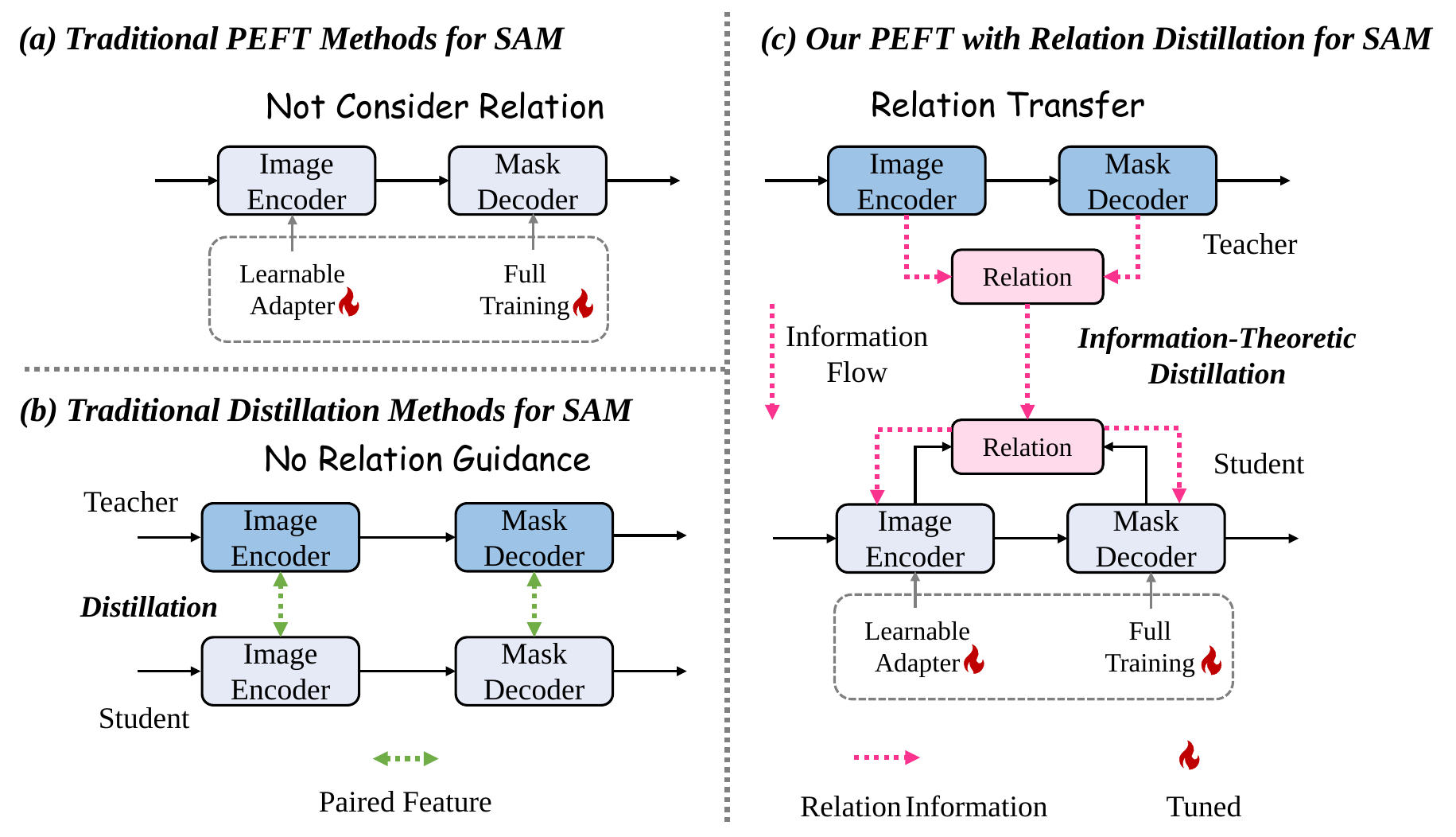}}
\caption{Comparing traditional PEFT and distillation paradigms with our proposed InfoSAM. (a) Existing PEFT methods for SAM directly adjust the trainable parameters of each module individually, often overlooking the cross-module relationships. (b) Traditional SAM distillation methods focus on model compression via paired feature alignment but lack relational guidance during projection training. (c) In contrast, our InfoSAM method enhances PEFT by incorporating information-theoretic distillation, enabling the transfer of domain-invariant relations from the pre-trained SAM to the fine-tuning student.}
\label{fig1_intro}
\end{center}
\vskip -0.2in
\end{figure}

Recently, the Segment Anything Model (SAM)~\cite{kirillov2023segment, ravi2024sam} emerged as a prominent foundation model for image segmentation.
While SAM demonstrates exceptional zero-shot performance on generic object segmentation, it often struggles with domain-specific real-world segmentation tasks~\cite{chen2023sam, zhong2024convolution}. Therefore, Parameter-Efficient Fine-Tuning (PEFT) for SAM~\cite{song2024simada,zhang2024blo,peng2024parameter} has gained attention as a promising solution, significantly reducing the fine-tuning costs associated with SAM's large pre-trained parameter set.
Existing PEFT methods for SAM primarily focus on fine-tuning the heavy image encoder~\cite{song2024simada,peng2024parameter} or aligning domain-specific features between the mask decoder and image encoder~\cite{xiao2025cat}. However, a promising improvement avenue is overlooked: preserving the beneficial information in pre-trained models.

Notably, SAM follows an encoder-decoder architecture, where the mask decoder refines the image embeddings extracted by the image encoder to localize objects. Unified training or fine-tuning methods~\cite{shu2025tinysam, xiao2025cat} have demonstrated effectiveness within this framework. This suggests that preserving the implicit relationship between the encoder and decoder could be beneficial for model fine-tuning. \textcolor{black}{This relationship may stem from extensive pre-training and be embedded in the feature distributions, making it delicate and easily disrupted by unrefined PEFT methods~\cite{wang2024samcl}. We argue that this is because task-specific tuning tends to override or suppress the universal visual features learned during pre-training.}


To enhance PEFT by leveraging implicit relationships, a natural approach is to extract these relationships from foundation models and inject them into fine-tuned models tailored for specific domains. However, not all implicit relationships are beneficial for downstream tasks—only the key domain-invariant relationships learned from across domains~\cite{hoffman2018cycada,xu2022dirl} contribute positively to every fine-tuned model. While knowledge distillation serves as a flexible bridge for transferring information between models~\cite{gou2021knowledge}, we propose to adopt a distillation approach between the pre-trained model and fine-tuned model to retain domain-invariant relationships.

Therefore, this brings us to two key challenges: \textit{1) How can we extract the domain-invariant relationship from pre-trained foundation models?} \textit{2) How can we effectively transfer the extracted information to fine-tuned models?} 



To address these challenges, we propose InfoSAM, a novel information-theoretical distillation method specifically designed for SAM PEFT. 
In order for the teacher to provide a good amount of the domain-invariant information, first we have to find out how this information could be quantified. 
To this aim, we introduce a robust and efficient Rényi's entropy-based quantification from information theory~\cite{ahn2019variational} to measure such a relation.
However, not all the relations in the pre-trained SAM are domain-invariant, there exists some pseudo-invariant information (e.g., color), which may negatively impact the generalization ability during the fine-tuning process~\cite{li2022invariant}. 
Therefore, \textbf{to address the first challenge}, we propose an attention-driven relation module specifically designed to extract critical structural patterns from the pre-trained SAM. By minimizing mutual information between the module's outputs and both encoder-decoder embeddings of SAM, it constructs an effective bottleneck that forces the module to maintain compressed yet domain-invariant representations.
Furthermore, \textbf{to tackle the second challenge}, we effectively distill the valuable relational knowledge from pre-trained SAM to the fine-tuned SAM by maximizing the mutual information between their extracted relations. This ensures faithful propagation of compressed semantic dependencies, thereby facilitating a more effective fine-tuning process. Our experiments on SAM and SAM2, evaluated across 4 diverse domains and 8 datasets, show that InfoSAM achieves superior adaptation and segmentation performance.

Overall, our contribution can be summarized as follows:
\begin{itemize}
    \item We present InfoSAM, the first information-theoretic framework for SAM adaptation, introducing an innovative distillation approach tailored for SAM PEFT to enhance performance in new scenarios.
    \item InfoSAM proposes novel dual complementary mechanisms for SAM adaptation: a relational bottleneck that strategically compresses task-irrelevant dependencies while preserving domain-invariant semantics, coupled with adaptive cross-model mutual information maximization ensuring provable preservation of essential structural knowledge.
    \item We conduct a comprehensive benchmark across diverse domains, including natural images, medical imaging, agriculture, and remote sensing. InfoSAM consistently demonstrates superior performance compared to other PEFT and distillation techniques across various downstream tasks.
\end{itemize}

\begin{figure*}[ht]
\vskip 0.2in
\begin{center}
\centerline{\includegraphics[width=1.85\columnwidth]{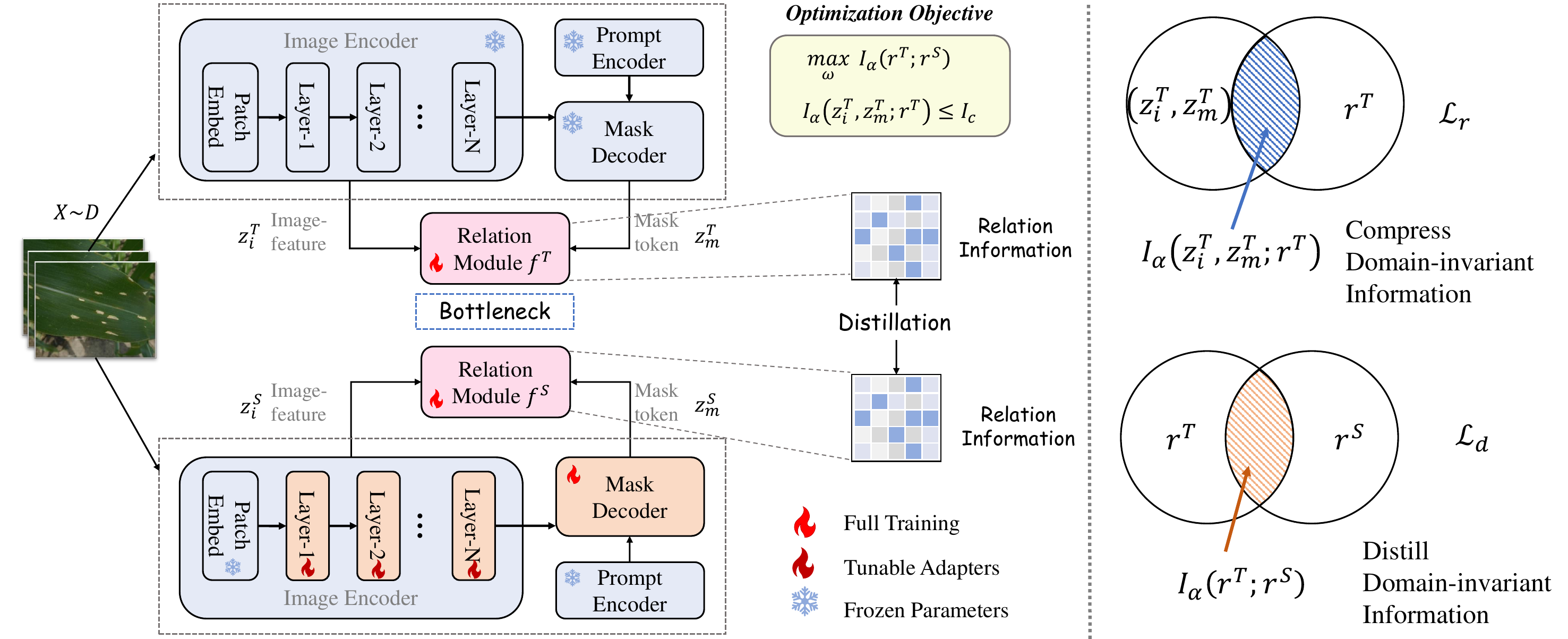}}
\caption{\textbf{The Flowchart of InfoSAM.} To leverage the domain-invariant relationships within modules from a well-trained foundation model (i.e., SAM) for enhancing PEFT. InfoSAM pioneers an information-theoretic framework for parameter-efficient SAM adaptation through two synergistic components: 1) Strategic compression of task-irrelevant dependencies while preserving domain-invariant feature relationships through optimized interaction between image embeddings $z^T_{i}$ and mask tokens $z^T_{m}$ (Eq.\ref{eq_compression}), and 2) Cross-model mutual information maximization to ensure faithful knowledge transfer (Eq.\ref{eq_6}). The right Venn diagrams illustrate the information constraint from the optimization problem.}
\label{fig2_overview}
\end{center}
\vskip -0.2in
\end{figure*}

\section{Related Work}
\label{related_work}


\subsection{Parameter Efficient Fine-Tuning for SAM}
\label{related_work:peft}
Parameter-Efficient Fine-Tuning (PEFT) alleviates the challenges of task-specific deployment in large foundation models by fine-tuning only a minimal subset of parameters while keeping the majority frozen.
Several prior works explore fine-tuning SAM for downstream tasks. SAM-Adapter~\cite{chen2023sam} is one of the pioneering works applying the PEFT method to SAM, incorporating task-specific prompts for each adapter. SU-SAM~\cite{song2024simada} presents a simple framework to efficiently fine-tune the SAM with Adapter or LoRA. SAM-COBOT~\cite{peng2024parameter} boosts existing PEFT techniques for fine-tuning SAM through cross-block orchestration. BLO-SAM~\cite{zhang2024blo} finetunes SAM based on bi-level optimization, eliminating the need for manual prompts by a learnable prompt embedding. Conv-LoRA~\cite{zhong2024convolution} integrates ultra-lightweight convolutional parameters into LoRA, injecting image-related inductive biases into the plain ViT encoder.

 \textcolor{black}{However, the above methods overlook preserving pre-trained information in foundation models during fine-tuning.} Our work explores enhancing fine-tuning methods for SAM from a novel perspective: leveraging information-based distillation to maintain domain-invariant relationships.

\subsection{Knowledge Distillation for SAM}
\label{related_work:kd}
Knowledge distillation (KD)~\cite{gou2021knowledge} effectively transfers knowledge from a large, well-trained model (teacher) to a smaller or simpler one (student). When applying KD to SAM, most efforts focus on compressing and transferring representations for downstream tasks. MobileSAM~\cite{zhang2023faster} distills SAM's ViT encoder into a TinyViT, while TinySAM~\cite{shu2025tinysam} uses full-stage KD. Other approaches distill SAM’s semantic priors for tasks like medical segmentation~\cite{dong2024efficient,shen2024fastsam3d} and image restoration~\cite{zhang2024distilling}. However, these methods focus on paired feature maps, neglecting the inter-module relationships within the teacher SAM. To address this, our approach utilizes information-theoretic principles to extract and transfer compact inter-module relationships to the student model.

\subsection{Domain-invariant Information in SAM}
 \textcolor{black}{The concept of domain-invariant information was first introduced in prior works on domain adaptive segmentation (DAS), which explored cross-domain invariant features such as edge and structural information~\cite{hoffman2018cycada}. DAS aims to learn domain-invariant representations across multiple domains and follows two main approaches: (i) extraction and refinement of domain-invariant features,where methods like feature disentanglement~\cite{chang2019all} or analysis~\cite{xu2022dirl} decompose images into domain-invariant (e.g., shapes, edges) and domain-specific (e.g., textures, colors) components, aiming to enhance the former while suppressing the latter; (2) GAN-based domain-invariant feature generation, which employs adversarial training to align domains at different levels: image~\cite{li2022stepwise}, feature~\cite{ma2024decomposition}, and output~\cite{huang2022multi}. For example, GLGAN~\cite{ma2024decomposition} integrates multi-scale global and local features to improve cross-domain transferability in remote sensing.}

 \textcolor{black}{SAM's large-scale pretraining encodes domain-invariant patterns for strong zero-shot generalization. Recent works leverage these universal visual patterns for downstream tasks~\cite{peng2024learning}. However, these methods rely on complex designs or external data to learn representations. In contrast, we focus on preserving the domain-invariant information in pre-trained SAM for fine-tuning.}




\section{Preliminaries}
\label{preliminaries}

\subsection{Rényi's $\alpha$-entropy and Mutual Information}
\label{pre_info}
In information theory, matrix-based Rényi's $\alpha$-entropy provides a novel way to quantify single-variable information or interactions across variables directly from samples. Unlike Shannon entropy, it leverages the eigenspectrum of a Gram matrix in reproducing kernel Hilbert space (RKHS), avoiding costly distribution evaluations\cite{gong2022computationally}.

\textbf{Definition 1.} Let $\kappa: \mathcal{X} \times \mathcal{X} \mapsto \mathbb{R}$ be an infinitely divisible positive kernel~\cite{bhatia2006infinitely}. Given $\{x_i \}_{i=1}^{n} \subset \mathcal{X}$, each $x_i$ being a real-valued scalar or vector, and the Gram matrix $K$ obtained from $K_{ij}=\kappa(x_i,x_j)$, a matrix-based analog to Rényi's entropy can be defined as:
\begin{equation}
\label{eq_eigs}
 \textcolor{black}{\mathbf{S}_\alpha(\mathbf{A})=\frac{1}{1-\alpha} \log _2\left[\sum_{i=1}^n \lambda_i^\alpha(\mathbf{A})\right]}
\end{equation}
where the kernel matrix $\mathbf{A}_{ij} = \frac{1}{n} \frac{K_{ij}}{\sqrt{K_{ii} K_{jj}}}$ is the normalized version of $K$ and $tr(\mathbf{A})=1$. The $\lambda_i(\mathbf{A})$ denotes the $i\-$th eigenvalue of $\mathbf{A}$.

\textbf{Definition 2.} Given $n$ pairs of samples $\left\{z_i=\left(x_i, y_i\right)\right\}_{i=1}^n$, and two positive definite kernels $\kappa_1: \mathcal{X} \times \mathcal{X} \mapsto \mathbb{R}$ and $\kappa_2: \mathcal{Y} \times \mathcal{Y} \mapsto \mathbb{R}$. After computing the Gram matrix $\mathbf{A}$ and $\mathbf{B}$, a joint Rényi's entropy can be defined as:
\begin{equation}
\label{eq_joint_entropy}
\mathbf{S}_\alpha(\mathbf{A}, \mathbf{B})=\mathbf{S}_\alpha\left(\frac{\mathbf{A} \circ \mathbf{B}}{\operatorname{tr}(\mathbf{A} \circ \mathbf{B})}\right)
\end{equation}
where $(\mathbf{A} \circ \mathbf{B})$ denotes the Hadamard product between the matrices $\mathbf{A}$ and $\mathbf{B}$. The mutual information $\mathbf{I}_\alpha(\mathbf{A} ; \mathbf{B})$ can be computed as:
\begin{equation}
\label{eq_mi}
\mathbf{I}_\alpha(\mathbf{A} ; \mathbf{B})=\mathbf{S}_\alpha(\mathbf{A})+\mathbf{S}_\alpha(\mathbf{B})-\mathbf{S}_\alpha(\mathbf{A}, \mathbf{B})
\end{equation}
The matrix-based Rényi's mutual information eliminates the need for high-dimensional probability density estimation of Shannon entropy, offering a more accurate and computationally efficient solution~\cite{dong2023optimal}.

\section{Methodology}
\label{methodology}

\subsection{Background and Notions}

The overview of InfoSAM is illustrated in Fig.~\ref{fig2_overview}. Given a teacher model and a student model, we denote the pre-trained SAM as $\phi^T$ and the fine-tuned SAM as $\phi^S$, which are parameterized by $\omega$.
Let $X \sim \mathcal{D}$ be an input sampled from the downstream dataset $\mathcal{D}$. The representations produced by $\phi^T(X)$ and $\phi^S(X)$ are defined as follows: The output features of the image encoder are denoted as $z_i^T$ and $z_i^S$, where $z_i^T, z_i^S \in \mathbb{R}^{B \times H \times W \times D}$. Here, $B$ is the batch size, $H$ and $W$ represent the height and width, respectively, and $D$ is the dimension of the image embeddings. Similarly, the output tokens from the two-way transformer in the mask decoder are denoted as $z_m^T$ and $z_m^S$, where $z_m^T, z_m^S \in \mathbb{R}^{B \times N \times D}$. These tokens encode the target mask information in a more abstract manner. Here, $N$ represents the number of masks, and the output token shares the same dimension $D$ as the image embeddings.

The goal of PEFT is to fine-tune $\phi^{S}(X; \omega)$ for adaptation to a new downstream task under the supervision of the teacher model $\phi^{T}$, where $\omega$ denotes the trainable PEFT parameters. To enhance the PEFT process using the frozen pre-trained teacher SAM, the loss can be formulated as:
\begin{equation}
\omega^* = \arg \min_{\omega} \mathcal{L}(X, Y \mid T),
\end{equation}
where $T$ represents intermediate features extracted by the teacher model, capturing the relational information within SAM modules, and $Y$ is the full dense label map. Following prior work, the task-specific loss function $\mathcal{L}(\cdot)$ is chosen as the structure loss~\cite{zhong2024convolution}.  

In this paper, rather than directly aligning paired representations between teacher and student~\cite{zhang2023faster,shu2025tinysam}, we leverage robust prior relational information from the pre-trained SAM to guide the PEFT process.

\subsection{An Information View of SAM Distillation}

\textbf{Problem Formulation.} Intuitively, the relationships between different modules in a well-trained foundation model are invaluable, as they are learned from extensive datasets. However, traditional PEFT methods for SAM, when fine-tuned to downstream tasks, risk disrupting these relationships. In this way, we need to address two key questions: how to capture critical relations from the pre-trained SAM and how to effectively transfer it to the fine-tuned model.

Firstly, by treating the teacher model as a mapping function, we argue that the critical relation information resides in multivariate mutual information $I_{\alpha}(z_{i}^T,z_{m}^T;r^T)$, where $z_{i}^T$, $z_{m}^T$ and $r^T$ represent the image embedding, mask token and relational interactions of the teacher, respectively. As such, $I_{\alpha}(z_{i}^T,z_{m}^T;r^T)$ quantifies how much information $r^T$ can tell about $(z_{i}^T,z_{m}^T)$.
Crucially, this mutual information constitutes a learnable bottleneck that fundamentally constrains the knowledge transfer process. The bottleneck mechanism enforces selective attention by restricting the information flow to a compressed representation, where only the most salient teacher-student interactions can be preserved. This is particularly vital as not all relations in the pre-trained SAM are universally transferable. For example, invariant features (e.g., geometric outlines) that exhibit cross-domain consistency are effective, while pseudo-invariant features (e.g., color distributions) that carry domain-specific biases need to be suppressed~\cite{li2022invariant}.

To prioritize domain-invariant relations, we constrain the information flow via an upper bound $I_c$:
\begin{equation}
\mathbf{I}_\alpha(z^{T}_{i}, z^{T}_{m}; \, r^{T}) \leq I_c
\label{eq_compression}
\end{equation}
where $ r^{T} = f^T(z^{T}_{i}, z^{T}_{m};\theta) $ represents the teacher’s relational mapping between image embeddings and mask tokens. The relation module is defined as  $f^T(z^{T}_{i}, z^{T}_{m};\theta): \mathbb{R}^{B \times H \times W \times D} \times \mathbb{R}^{B \times N \times D} \rightarrow \mathbb{R}^{B \times N \times (H \cdot W)}$. The $\theta$ represents the parameters of the relation module. This compression forces the module to retain only essential information for distillation.

After that, we employ a distillation approach to transfer the core relationships by maximizing their mutual information:
\begin{equation}
\begin{aligned}
\underset{\omega}{\max} \quad & \mathbf{I}_\alpha(r^T; r^S) \\
\text{subject to} \quad & \mathbf{I}_\alpha(z^{T}_{i}, z^{T}_{m}; \, r^{T}) \leq I_c
\label{eq_6}
\end{aligned}
\end{equation}
where $r^S=f^S(z^{S}_{i}, z^{S}_{m};\theta)$ denotes the student's relation in fine-tuned SAM, with $f^S$ sharing the same parameters as $f^T$. The information bottleneck principle operates through two coupled mechanisms: 
(i) \textit{compression} via minimizing $\mathbf{I}_\alpha(z^T_i,z^T_m;r^T)$ to extract minimal sufficient statistics $r^T$ from $(z^T_i,z^T_m)$, and 
(ii) \textit{distillation} via maximizing $\mathbf{I}_\alpha(r^T;r^S)$ to preserve maximal predictive information. The Lagrangian formulation explicitly implements this trade-off:
\begin{equation}
\begin{aligned}
\underset{\omega}{\max} \quad & \mathbf{I}_\alpha(r^T; r^S) - \beta\mathbf{I}_\alpha(z^{T}_{i}, z^{T}_{m}; \, r^{T})
\end{aligned}
\end{equation}
where $\beta$ is a hyper-parameter for trade-off. 



\begin{figure}[t]
\vskip 0.2in
\begin{center}
\centerline{\includegraphics[width=\columnwidth]{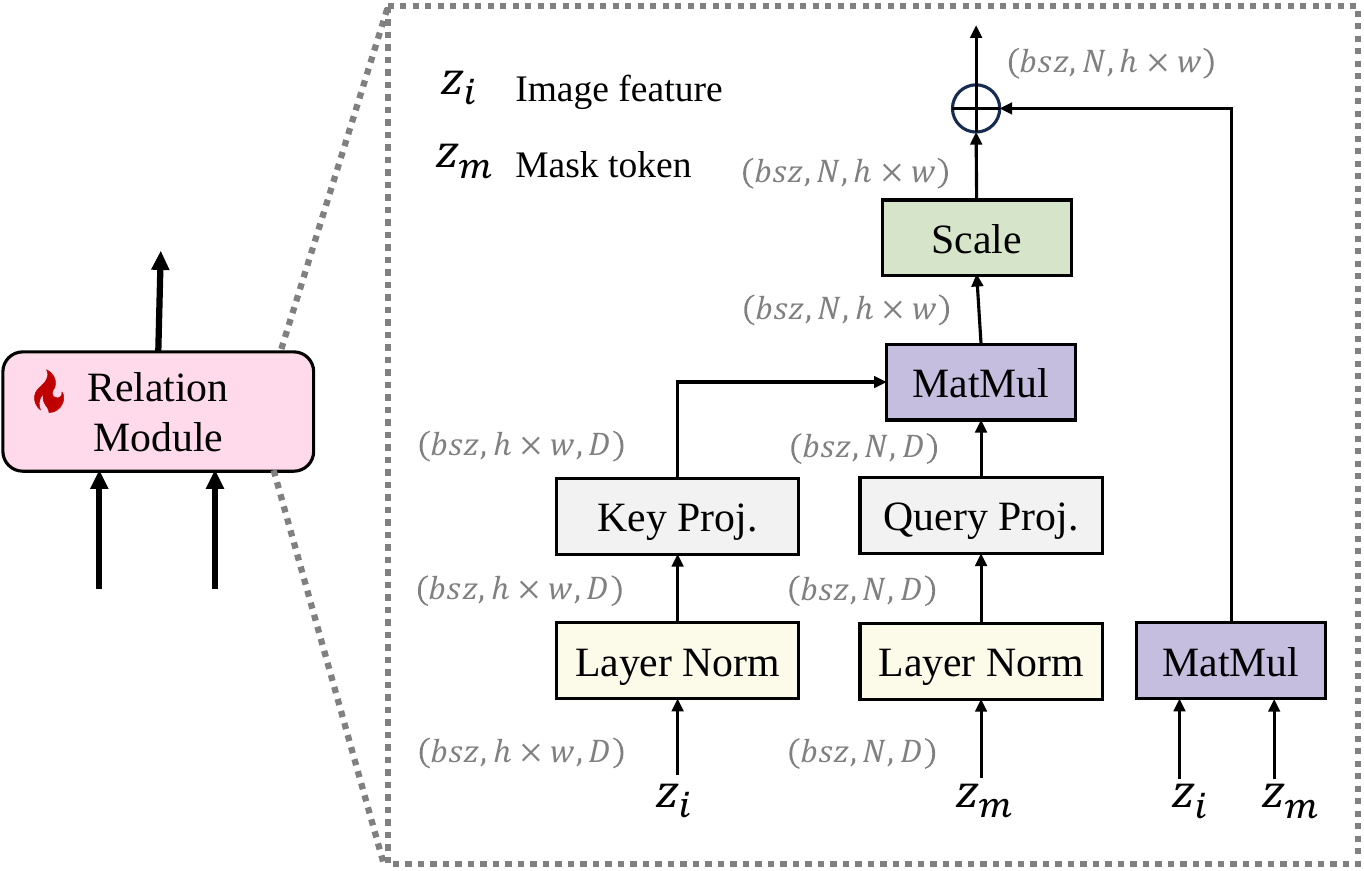}}
\caption{The architecture of attention-based relation module. It is designed to capture the relationship between image encoder and mask decoder, facilitating effective interaction between these components in SAM.}
\label{fig3_rm}
\end{center}
\vskip -0.2in
\end{figure}

\textbf{Compressing Intra-SAM Relations.} To efficiently capture the relationship within pre-trained SAM, we propose an attention-based module designed for extraction, illustrated in Fig.~\ref{fig3_rm}. Given $z^{T}_{i}$, the output of the image encoder of SAM, and $z^{T}_{m}$, the mask token embedding, as the input of relation module $f^T$. It mainly uses a combination of attention mechanisms and residual connections. 

First, both $z^{T}_{i}$ and $z^{T}_{m}$ are passed through a Layer Normalization step to stabilize the features. After that, $z^{T}_{m}$ and $z^{T}_{i}$ are linearly projected into a query vector $ Q \in \mathbb{R}^{B \times N \times D}$ and a key vector $K \in \mathbb{R}^{B \times (H \cdot W) \times D}$, respectively:
\begin{equation}
\begin{aligned}
Q = W_Q \cdot LayerNorm(z^T_m), \\
\quad K = W_K \cdot LayerNorm(z^T_i)
\end{aligned}
\end{equation}
where $W_Q, W_K \in \mathbb{R}^{D \times D} $ are learnable projection matrices. The attention scores are computed by combining two components: the scaled dot product of $Q$ and $K$ and the residuals from the dot product of $z^T_m$ and $z^T_i$. The scores are summed to produce the final attention map:
\begin{equation}
\begin{aligned}
S_\alpha = \frac{Q K^\top}{\sqrt{D}} + z^T_m \cdot z^{T \top}_i
\end{aligned}
\end{equation}
where $S_\alpha$ is the attention score. To ensure consistency and comparability, $\alpha$ is flattened and normalized using $\ell_2$-normalization, resulting in the final output of $f^T$, denoted as $r^T$. To encourage the relation encoding process to focus on domain-invariant information, the first loss for relation compression can be expressed as:
\begin{equation}
\begin{aligned}
\label{eq_loss_r_mi}
\mathcal{L}_r = &\mathbf{I}_\alpha(z^{T}_{i}, z^{T}_{m};r^{T}) \\
       = &\cancel{\mathbf{S}_\alpha(G^{T}_{i}, G^{T}_{m})}+\mathbf{S}_\alpha(G^{T}_r)-\mathbf{S}_\alpha(G^{T}_{i}, G^{T}_{m},G^{T}_r)
\end{aligned}
\end{equation}
where $G^{T}_{i}, G^{T}_{m}, G^{T}_r \in \mathbb{R}^{N \times N}$ are the Gram matrices induced by a batch of normalized features $z^{T}_{i}$, $z^{T}_{m}$, and the output of $r^{T}$ with a polynomial kernel of degree 1. Notably, the teacher entropy term in this loss is excluded, as the teacher's weights remain fixed during PEFT. 

According to Eq.(\ref{eq_eigs}), computing eigenvalues of large matrices is computationally intensive ~\cite{kerr2009qr, yu2019multivariate}. To mitigate this, we set $\alpha = 2$, allowing us to compute matrix-based Rényi's $\alpha$-entropy via the Frobenius norm: \textcolor{black}{$|\mathbf{A}\|_F^2 = \mathrm{tr}(\mathbf{A}\mathbf{A}^H) = \sum_{i=1}^n \lambda_i^2(\mathbf{A})$}. Consequently, $L_r$ can be reformulated as:
\begin{equation}
\begin{aligned}
\label{eq_lossr}
\mathcal{L}_r = - \log_2 \| G^{T}_{r} \|_F^2 + \log_2 \| G^{T}_{imr} \|_F^2
\end{aligned}
\end{equation}
where $G^{T}_{imr}=G^{T}_{i} \circ G^{T}_{m} \circ G^{T}_{r}$. The $\circ$ is Hadamard product. The first term in $\mathcal{L}_r$ acts as a spectral compression regularizer that constrains the relation module and encourages it to learn more compact and refined representations. The second term minimizes the joint entropy of the feature interactions across the image encoder, mask decoder, and relation module, effectively filtering spurious relationships and preserving domain-invariant interactions critical for cross-domain adaptation. 

\textbf{Maximizing Inter-SAM Relations.} 
After extracting the essential relationships between the image encoder and the mask decoder, we transfer the relationships by minimizing their distance. A natural choice to accomplish this is by maximizing the mutual information between the two representations. While most existing works~\cite{ahn2019variational, kuang2023improving} focus on minimizing a lower bound of mutual information, we directly maximize the matrix-based Rényi's mutual information itself to avoid the expensive evaluation of underlying distribution for distillation loss:
\begin{equation}
\begin{aligned}
\label{eq_loss_d}
\mathcal{L}_d = & -\mathbf{I}_\alpha(r^{T}; r^{S}) \\
       = &-\mathbf{S}_\alpha(G^{T}_r) - \mathbf{S}_\alpha(G^{S}_r)+\mathbf{S}_\alpha(G^{T}_{r}, G^{S}_{r})
\end{aligned}
\end{equation}
Similarly, $G^{T}_r$ and  $G^{S}_r$ denote the Gram matrices corresponding to the student and teacher relations, respectively.
We denote the $G_r^{TS}= G^{T}_{r} \circ G^{S}_{r} $, then the distillation loss can be expressed as:
\begin{equation}
\begin{aligned}
\label{eq_lossd}
\mathcal{L}_d = \log_2 \| G^{T}_{r} \|_F^2 + \log_2 \| G^{S}_{r} \|_F^2 - \log_2 \| G^{TS}_{r} \|_F^2
\end{aligned}
\end{equation}
Consistent with $\mathcal{L}_r$, the $\mathcal{L}_d$ also sets the entropy order $\alpha$ to 2 and utilizes the Frobenius norm for equivalent transformation. From this perspective, the components of  $\mathcal{L}_d$ can be viewed as regularization terms (i.e., the first two terms) and a relation alignment (i.e., the third term) between two models, while $\log_2$ improving robustness to relations.

Overall, combining Eq.(\ref{eq_lossr}) and Eq.(\ref{eq_lossd}), the final objective of relation compression and transfer can be defined as:
\begin{equation}
\begin{aligned}
\label{eq_loss_info}
 \textcolor{black}{\mathcal{L}_{info} = \lambda_1 * \mathcal{L}_{r} + \lambda_2 * \mathcal{L}_{d}}
\end{aligned}
\end{equation}
where $\lambda_1$ and $\lambda_2$ are hyper-parameters to trade-off between sufficiency (domain-invariant information transmitted from $r^T$ to $r^S$) and minimality (the complexity of $r^T$). Further details and PyTorch-style pseudocode for InfoSAM are provided in Appendix \ref{app_derive_loss} and \ref{pseudocode}.

\begin{table*}[ht]
\caption{\textbf{Comparison of PEFT methods for SAM across various downstream segmentation tasks.} All results are based on the ViT-B backbone. ``SAM": without adaptation. ``decoder-only": directly fine-tuning the mask decoder of SAM.}
\label{tab:peft_comparison}
\begin{center}
\begin{scriptsize}
\setlength{\tabcolsep}{4pt}
\begin{tabular}{@{}l|ccc|cccc|cc|cc@{}}
\toprule
\multirow{3}{*}{\textsc{\textbf{Method}}} & \multicolumn{3}{c|}{\textsc{\textbf{Natural Images}}}  & \multicolumn{4}{c|}{\textsc{\textbf{Medical}}} & \multicolumn{2}{c|}{\textsc{\textbf{Agriculture}}} & \multicolumn{2}{c}{\textsc{\textbf{Remote Sensing}}} \\ \cmidrule(l){2-12} 
                                 & \multicolumn{3}{c|}{CAMO}                                                                        & \multicolumn{2}{c|}{ISIC 2017}                                                       & \multicolumn{2}{c|}{Kvasir}                                                                              & \multicolumn{2}{c|}{Leaf}                                                         & \multicolumn{2}{c}{Road}                                                          \\
                                 & $S_{\alpha} \uparrow$                   & $E_{\phi} \uparrow$                     & $F_{\beta}^{\omega} \uparrow$           & Jac $\uparrow$                           & \multicolumn{1}{c|}{Dice $\uparrow$}                           & $S_{\alpha} \uparrow$                                       & $E_{\phi} \uparrow$                                          & IoU $\uparrow $                                    & Dice $\uparrow$                                  & IoU  $\uparrow$                                   & Dice $\uparrow$                                   \\ \midrule
SAM                              & 79.7\textsubscript{$\pm$ 0.02}          & 88.8\textsubscript{$\pm$ 0.09}          & 79.6\textsubscript{$\pm$ 0.01}          & 61.0\textsubscript{$\pm$ 0.12}          & \multicolumn{1}{c|}{71.7\textsubscript{$\pm$ 0.14}}          & 71.4\textsubscript{$\pm$ 0.16}          & 77.9\textsubscript{$\pm$ 0.17}          & 37.6\textsubscript{$\pm$ 0.11}          & 47.0\textsubscript{$\pm$ 0.16}          & 7.2\textsubscript{$\pm$ 0.24}           & 12.9\textsubscript{$\pm$ 0.29}          \\
decoder-only                     & 84.9\textsubscript{$\pm$ 0.38}          & 92.7\textsubscript{$\pm$ 0.34}          & 81.8\textsubscript{$\pm$ 0.33}          & 85.9\textsubscript{$\pm$ 0.34}          & \multicolumn{1}{c|}{92.2\textsubscript{$\pm$ 0.20}}          & 90.9\textsubscript{$\pm$ 0.05}          & 95.2\textsubscript{$\pm$ 0.18}          & 55.6\textsubscript{$\pm$ 1.12}          & 68.8\textsubscript{$\pm$ 1.17}          & 47.6\textsubscript{$\pm$ 0.47}          & 64.1\textsubscript{$\pm$ 0.47}          \\ \midrule
BitFit                           & 87.5\textsubscript{$\pm$ 0.13}          & 94.5\textsubscript{$\pm$ 0.08}          & 85.3\textsubscript{$\pm$ 0.48}          & 87.7\textsubscript{$\pm$ 0.14}          & \multicolumn{1}{c|}{93.2\textsubscript{$\pm$ 0.08}}          & 92.5\textsubscript{$\pm$ 0.12}          & 96.3\textsubscript{$\pm$ 0.20}          & 69.2\textsubscript{$\pm$ 0.67}          & 80.3\textsubscript{$\pm$ 0.68}          & 58.1\textsubscript{$\pm$ 0.06}          & 73.1\textsubscript{$\pm$ 0.06}          \\
AdaptFormer                      & 87.9\textsubscript{$\pm$ 0.10}          & 94.8\textsubscript{$\pm$ 0.21}          & 86.2\textsubscript{$\pm$ 0.19}          & 87.6\textsubscript{$\pm$ 0.24}          & \multicolumn{1}{c|}{93.2\textsubscript{$\pm$ 0.15}}          & 93.3\textsubscript{$\pm$ 0.68}          & 97.0\textsubscript{$\pm$ 0.81}          & 75.0\textsubscript{$\pm$ 0.11}          & 84.8\textsubscript{$\pm$ 0.08}          & 61.1\textsubscript{$\pm$ 0.15}          & 75.5\textsubscript{$\pm$ 0.12}          \\
LoRA                             & 87.7\textsubscript{$\pm$ 0.59}          & 94.6\textsubscript{$\pm$ 0.50}          & 85.1\textsubscript{$\pm$ 0.64}          & 87.8\textsubscript{$\pm$ 0.24}          & \multicolumn{1}{c|}{93.3\textsubscript{$\pm$ 0.13}}          & 93.0\textsubscript{$\pm$ 0.14}          & 96.6\textsubscript{$\pm$ 0.11}          & 71.4\textsubscript{$\pm$ 0.54}          & 82.1\textsubscript{$\pm$ 0.62}          & 59.0\textsubscript{$\pm$ 0.19}          & 74.0\textsubscript{$\pm$ 0.17}          \\
Adapter                          & 88.2\textsubscript{$\pm$ 0.44}          & 94.8\textsubscript{$\pm$ 0.34}          & 86.7\textsubscript{$\pm$ 0.92}          & 87.7\textsubscript{$\pm$ 0.23}          & \multicolumn{1}{c|}{93.2\textsubscript{$\pm$ 0.16}}          & 93.4\textsubscript{$\pm$ 0.12}          & 97.1\textsubscript{$\pm$ 0.15}          & 74.4\textsubscript{$\pm$ 0.16}          & 84.3\textsubscript{$\pm$ 0.28}          & 60.5\textsubscript{$\pm$ 0.10}          & 75.1\textsubscript{$\pm$ 0.08}          \\ \midrule
HQ-SAM                           & 85.1\textsubscript{$\pm$ 0.10}          & 92.6\textsubscript{$\pm$ 0.10}          & 81.0\textsubscript{$\pm$ 0.61}          & 86.3\textsubscript{$\pm$ 0.32}          & \multicolumn{1}{c|}{92.4\textsubscript{$\pm$ 0.19}}          & 91.1\textsubscript{$\pm$ 0.50}          & 95.5\textsubscript{$\pm$ 0.57}          & 66.2\textsubscript{$\pm$ 0.44}          & 77.8\textsubscript{$\pm$ 0.43}          & 54.9\textsubscript{$\pm$ 0.16}          & 70.6\textsubscript{$\pm$ 0.13}          \\
SU-SAM                           & 88.3\textsubscript{$\pm$ 0.21}          & 95.0\textsubscript{$\pm$ 0.22}          & 86.2\textsubscript{$\pm$ 0.59}          & 87.8\textsubscript{$\pm$ 0.18}          & \multicolumn{1}{c|}{93.2\textsubscript{$\pm$ 0.09}}          & 93.8\textsubscript{$\pm$ 0.02}          & 97.5\textsubscript{$\pm$ 0.06}          & 74.7\textsubscript{$\pm$ 0.53}          & 84.5\textsubscript{$\pm$ 0.56}          & 60.2\textsubscript{$\pm$ 0.26}          & 74.8\textsubscript{$\pm$ 0.22}          \\
ConvLoRA-SAM                     & 87.5\textsubscript{$\pm$ 0.39}          & 94.5\textsubscript{$\pm$ 0.17}          & 85.4\textsubscript{$\pm$ 0.41}          & 87.7\textsubscript{$\pm$ 0.22}          & \multicolumn{1}{c|}{93.2\textsubscript{$\pm$ 0.11}}          & 92.9\textsubscript{$\pm$ 0.13}          & 96.6\textsubscript{$\pm$ 0.28}          & 71.4\textsubscript{$\pm$ 0.44}          & 82.2\textsubscript{$\pm$ 0.37}          & 59.6\textsubscript{$\pm$ 0.22}          & 74.4\textsubscript{$\pm$ 0.20}          \\ \midrule
\textbf{LoRA+Ours}               & \textbf{88.3\textsubscript{$\pm$ 0.05}} & \textbf{95.2\textsubscript{$\pm$ 0.00}} & \textbf{85.8\textsubscript{$\pm$ 0.59}} & \textbf{88.1\textsubscript{$\pm$ 0.08}} & \multicolumn{1}{c|}{\textbf{93.5\textsubscript{$\pm$ 0.05}}} & \textbf{93.4\textsubscript{$\pm$ 0.11}} & \textbf{96.8\textsubscript{$\pm$ 0.09}} & \textbf{72.2\textsubscript{$\pm$ 0.06}} & \textbf{82.8\textsubscript{$\pm$ 0.04}} & \textbf{59.9\textsubscript{$\pm$ 0.20}} & \textbf{74.6\textsubscript{$\pm$ 0.17}} \\
\textbf{Adapter+Ours}            & \textbf{88.6\textsubscript{$\pm$ 0.09}} & \textbf{95.1\textsubscript{$\pm$ 0.05}} & \textbf{87.1\textsubscript{$\pm$ 0.37}} & \textbf{88.0\textsubscript{$\pm$ 0.05}} & \multicolumn{1}{c|}{\textbf{93.4\textsubscript{$\pm$ 0.00}}} & \textbf{94.4\textsubscript{$\pm$ 0.12}} & \textbf{97.9\textsubscript{$\pm$ 0.09}} & \textbf{75.6\textsubscript{$\pm$ 0.27}} & \textbf{85.2\textsubscript{$\pm$ 0.23}} & \textbf{61.4\textsubscript{$\pm$ 0.30}} & \textbf{75.8\textsubscript{$\pm$ 0.27}} \\ \bottomrule
\end{tabular}
\end{scriptsize}
\end{center}
\end{table*}

\begin{table*}[ht]
\caption{\textbf{Comparison of distillation methods for SAM fine-tuning across various domains. } ``Teacher": SAM without adaptation. ``Student":  fine-tune SAM's image encoder by adding adapters, while fully training the decoder directly. All compared methods utilize student models with the same adapter-based structure.}
\label{tab:distillation_comparison}
\begin{center}
\begin{scriptsize}
\setlength{\tabcolsep}{4pt}
\begin{tabular}{@{}l|ccc|cccc|cc|cc@{}}
\toprule
\multirow{3}{*}{\textsc{\textbf{Method}}} & \multicolumn{3}{c|}{\textsc{\textbf{Natural Images}}}  & \multicolumn{4}{c|}{\textsc{\textbf{Medical}}} & \multicolumn{2}{c|}{\textsc{\textbf{Agriculture}}} & \multicolumn{2}{c}{\textsc{\textbf{Remote Sensing}}} \\ \cmidrule(l){2-12} 
                                 & \multicolumn{3}{c|}{CAMO}                                                                        & \multicolumn{2}{c|}{ISIC 2017}                                                       & \multicolumn{2}{c|}{Kvasir}                                                                              & \multicolumn{2}{c|}{Leaf}                                                         & \multicolumn{2}{c}{Road}                                                          \\
                                 & $S_{\alpha} \uparrow$                   & $E_{\phi} \uparrow$                     & $F_{\beta}^{\omega} \uparrow$           & Jac $\uparrow$                           & \multicolumn{1}{c|}{Dice $\uparrow$}                           & $S_{\alpha} \uparrow$                                       & $E_{\phi} \uparrow$                                          & IoU $\uparrow $                                    & Dice $\uparrow$                                  & IoU  $\uparrow$                                   & Dice $\uparrow$                                    \\ \midrule
Teacher                          & 79.7\textsubscript{$\pm$ 0.02}          & 88.8\textsubscript{$\pm$ 0.09}          & 79.6\textsubscript{$\pm$ 0.01}          & 61.0\textsubscript{$\pm$ 0.12}          & \multicolumn{1}{c|}{71.7\textsubscript{$\pm$ 0.14}}          & 83.0\textsubscript{$\pm$ 0.10}          & 88.8\textsubscript{$\pm$ 0.29}          & 37.6\textsubscript{$\pm$ 0.11}          & 47.0\textsubscript{$\pm$ 0.16}          & 7.2\textsubscript{$\pm$ 0.24}           & 12.9\textsubscript{$\pm$ 0.29}          \\
Student                          & 88.2\textsubscript{$\pm$ 0.44}          & 94.8\textsubscript{$\pm$ 0.34}          & 86.7\textsubscript{$\pm$ 0.92}          & 87.7\textsubscript{$\pm$ 0.23}          & \multicolumn{1}{c|}{93.2\textsubscript{$\pm$ 0.16}}          & 93.4\textsubscript{$\pm$ 0.12}          & 97.1\textsubscript{$\pm$ 0.15}          & 74.4\textsubscript{$\pm$ 0.16}          & 84.3\textsubscript{$\pm$ 0.28}          & 60.5\textsubscript{$\pm$ 0.10}          & 75.1\textsubscript{$\pm$ 0.08}          \\ \midrule
Logit                            & 88.4\textsubscript{$\pm$ 0.08}          & 94.9\textsubscript{$\pm$ 0.05}          & 87.1\textsubscript{$\pm$ 0.22}          & 87.2\textsubscript{$\pm$ 0.43}          & \multicolumn{1}{c|}{92.9\textsubscript{$\pm$ 0.29}}          & 93.2\textsubscript{$\pm$ 0.19}          & 96.5\textsubscript{$\pm$ 0.19}          & 73.0\textsubscript{$\pm$ 0.35}          & 83.3\textsubscript{$\pm$ 0.29}          & 50.9\textsubscript{$\pm$ 0.08}          & 67.2\textsubscript{$\pm$ 0.06}          \\ \midrule
PKD                              & 87.0\textsubscript{$\pm$ 0.43}          & 94.1\textsubscript{$\pm$ 0.23}          & 84.3\textsubscript{$\pm$ 0.97}          & 86.5\textsubscript{$\pm$ 0.26}          & \multicolumn{1}{c|}{92.5\textsubscript{$\pm$ 0.17}}          & 92.2\textsubscript{$\pm$ 0.25}          & 96.0\textsubscript{$\pm$ 0.17}          & 70.2\textsubscript{$\pm$ 1.15}          & 81.1\textsubscript{$\pm$ 1.08}          & 56.9\textsubscript{$\pm$ 0.61}          & 72.2\textsubscript{$\pm$ 0.56}          \\
PKT                              & 87.8\textsubscript{$\pm$ 0.40}          & 94.5\textsubscript{$\pm$ 0.35}          & 86.2\textsubscript{$\pm$ 0.46}          & 87.4\textsubscript{$\pm$ 0.12}          & \multicolumn{1}{c|}{93.0\textsubscript{$\pm$ 0.07}}          & 93.7\textsubscript{$\pm$ 0.41}          & 97.3\textsubscript{$\pm$ 0.53}          & 74.2\textsubscript{$\pm$ 0.51}          & 84.2\textsubscript{$\pm$ 0.52}          & 60.7\textsubscript{$\pm$ 0.20}          & 75.2\textsubscript{$\pm$ 0.16}          \\ \midrule
IBD                              & 85.2\textsubscript{$\pm$ 0.47}          & 92.6\textsubscript{$\pm$ 0.35}          & 82.4\textsubscript{$\pm$ 0.31}          & 85.1\textsubscript{$\pm$ 0.74}          & \multicolumn{1}{c|}{91.7\textsubscript{$\pm$ 0.45}}          & 91.5\textsubscript{$\pm$ 0.14}          & 95.3\textsubscript{$\pm$ 0.05}          & 72.2\textsubscript{$\pm$ 0.12}          & 82.7\textsubscript{$\pm$ 0.07}          & 44.9\textsubscript{$\pm$ 0.18}          & 61.5\textsubscript{$\pm$ 0.18}          \\
VID                              & 87.9\textsubscript{$\pm$ 0.22}          & 94.8\textsubscript{$\pm$ 0.34}          & 86.3\textsubscript{$\pm$ 0.32}          & 87.6\textsubscript{$\pm$ 0.44}          & \multicolumn{1}{c|}{93.1\textsubscript{$\pm$ 0.29}}          & 93.7\textsubscript{$\pm$ 0.16}          & 97.4\textsubscript{$\pm$ 0.07}          & 75.1\textsubscript{$\pm$ 0.08}          & 84.9\textsubscript{$\pm$ 0.17}          & 60.7\textsubscript{$\pm$ 0.19}          & 75.4\textsubscript{$\pm$ 0.19}          \\ \midrule
SemCKD                           & 86.2\textsubscript{$\pm$ 0.16}          & 93.5\textsubscript{$\pm$ 0.21}          & 82.8\textsubscript{$\pm$ 1.54}          & 85.4\textsubscript{$\pm$ 0.27}          & \multicolumn{1}{c|}{91.8\textsubscript{$\pm$ 0.19}}          & 92.4\textsubscript{$\pm$ 0.07}          & 96.2\textsubscript{$\pm$ 0.03}          & 72.0\textsubscript{$\pm$ 0.04}          & 82.8\textsubscript{$\pm$ 0.10}          & 53.5\textsubscript{$\pm$ 0.17}          & 69.4\textsubscript{$\pm$ 0.17}          \\
ReviewKD                         & 86.7\textsubscript{$\pm$ 0.07}          & 94.0\textsubscript{$\pm$ 0.09}          & 84.6\textsubscript{$\pm$ 0.63}          & 85.5\textsubscript{$\pm$ 0.26}          & \multicolumn{1}{c|}{91.9\textsubscript{$\pm$ 0.15}}          & 92.4\textsubscript{$\pm$ 0.33}          & 96.4\textsubscript{$\pm$ 0.26}          & 72.6\textsubscript{$\pm$ 0.64}          & 83.1\textsubscript{$\pm$ 0.47}          & 57.3\textsubscript{$\pm$ 0.11}          & 72.6\textsubscript{$\pm$ 0.11}          \\ \midrule
TinySAM                          & 83.7\textsubscript{$\pm$ 0.39}          & 91.6\textsubscript{$\pm$ 0.31}          & 81.1\textsubscript{$\pm$ 0.35}          & 79.4\textsubscript{$\pm$ 1.12}          & \multicolumn{1}{c|}{87.8\textsubscript{$\pm$ 0.84}}          & 88.5\textsubscript{$\pm$ 0.31}          & 93.5\textsubscript{$\pm$ 0.24}          & 48.6\textsubscript{$\pm$ 1.14}          & 61.0\textsubscript{$\pm$ 0.95}          & 25.7\textsubscript{$\pm$ 1.19}          & 39.6\textsubscript{$\pm$ 1.71}          \\
MobileSAM                        & 87.1\textsubscript{$\pm$ 0.36}          & 94.1\textsubscript{$\pm$ 0.27}          & 85.1\textsubscript{$\pm$ 0.09}          & 86.7\textsubscript{$\pm$ 0.13}          & \multicolumn{1}{c|}{92.6\textsubscript{$\pm$ 0.09}}          & 92.5\textsubscript{$\pm$ 0.12}          & 96.3\textsubscript{$\pm$ 0.14}          & 71.9\textsubscript{$\pm$ 0.30}          & 82.6\textsubscript{$\pm$ 0.39}          & 59.2\textsubscript{$\pm$ 0.09}          & 74.1\textsubscript{$\pm$ 0.08}          \\
\textbf{InfoSAM(Ours)}           & \textbf{88.6\textsubscript{$\pm$ 0.09}} & \textbf{95.1\textsubscript{$\pm$ 0.05}} & \textbf{87.1\textsubscript{$\pm$ 0.37}} & \textbf{88.0\textsubscript{$\pm$ 0.05}} & \multicolumn{1}{c|}{\textbf{93.4\textsubscript{$\pm$ 0.00}}} & \textbf{94.4\textsubscript{$\pm$ 0.12}} & \textbf{97.9\textsubscript{$\pm$ 0.09}} & \textbf{75.6\textsubscript{$\pm$ 0.27}} & \textbf{85.2\textsubscript{$\pm$ 0.23}} & \textbf{61.4\textsubscript{$\pm$ 0.30}} & \textbf{75.8\textsubscript{$\pm$ 0.27}} \\ \bottomrule
\end{tabular}
\end{scriptsize}
\end{center}
\end{table*}

\subsection{Applying information theory to SAM}

\textbf{Overall Loss Function.} Following previous works~\cite{zhong2024convolution}, we incorporate the proposed information-theoretic distillation loss $\mathcal{L}_{info}$ with a structure loss $\mathcal{L}_{ce}$~\cite{fan2020pranet}, which combines the weighted IoU loss and binary cross-entropy loss. The overall loss function is derived as:
\begin{equation}
\begin{aligned}
\label{eq_loss_all}
\mathcal{L} = \mathcal{L}_{ce} + \mathcal{L}_{info}
\end{aligned}
\end{equation}
Finally, we employ this new loss function for fine-tuning SAM. During fine-tuning, we first learn robust relations and then transfer this knowledge. The $\mathcal{L}_{info}$ regulates information flow between SAM's hierarchical representations, avoiding over-retention of low-level details while enhancing geometrically critical features. This aligns with the rate-distortion tradeoff in information bottleneck theory~\cite{tishby2015deep}, where information is compressed and then generalized.

\begin{table*}[t]
\caption{\textbf{Comparison of PEFT methods and distillation methods with SAM2 across various domains.} All results are based on the Hiera-B+ backbone. ``SAM2": without adaptation.}
\vskip -0.1in
\label{tab:peft_and_distill_comparison_sam2}
\begin{center}
\begin{scriptsize}
\setlength{\tabcolsep}{4pt}
\begin{tabular*}{\textwidth}{@{}c@{}c@{}}
\begin{minipage}[t]{0.48\textwidth}
\centering
\setlength{\tabcolsep}{4pt}
\caption*{(a) PEFT Methods Comparison}
\begin{tabular}{@{}lccc@{}}
\toprule
\multirow{2}{*}{\textsc{\textbf{Method}}} & \textsc{\textbf{Medical}} & \textsc{\textbf{Agriculture}} & \textsc{\textbf{Remote Sensing}}  \\ \cmidrule(l){2-4} 
                                 & $S_{\alpha}$  (Kvasir)                  & IoU  (Leaf)                         & IoU  (Road)                          \\ \midrule
SAM2                             & 87.1\textsubscript{$\pm$ 0.12} & 42.7\textsubscript{$\pm$ 0.32} & 6.9\textsubscript{$\pm$ 0.13} \\
decoder-only                     & 93.2\textsubscript{$\pm$ 0.07} & 71.8\textsubscript{$\pm$ 0.58} & 48.5\textsubscript{$\pm$ 0.47} \\\midrule
BitFit                           & 93.8\textsubscript{$\pm$ 0.09} & 75.4\textsubscript{$\pm$ 0.29} & 59.2\textsubscript{$\pm$ 0.26} \\
AdaptFormer                      & 93.7\textsubscript{$\pm$ 0.19} & 73.6\textsubscript{$\pm$ 1.10} & 59.9\textsubscript{$\pm$ 0.35} \\
LoRA                             & 93.7\textsubscript{$\pm$ 0.10} & 75.9\textsubscript{$\pm$ 0.40} & 60.8\textsubscript{$\pm$ 0.32} \\
Adapter                          & 94.4\textsubscript{$\pm$ 0.06} & 76.8\textsubscript{$\pm$ 0.56} & 60.9\textsubscript{$\pm$ 0.14} \\\midrule
\textbf{LoRA+Ours}               & \textbf{94.0\textsubscript{$\pm$ 0.09}} & \textbf{76.1\textsubscript{$\pm$ 0.38}} & \textbf{60.9\textsubscript{$\pm$ 0.05}} \\
\textbf{Adapter+Ours}            & \textbf{94.5\textsubscript{$\pm$ 0.17}} & \textbf{77.3\textsubscript{$\pm$ 0.14}} & \textbf{61.3\textsubscript{$\pm$ 0.05}} \\
\bottomrule
\end{tabular}
\end{minipage} &
\begin{minipage}[t]{0.48\textwidth}
\centering
\setlength{\tabcolsep}{4pt}
\caption*{(b) Distillation Methods Comparison}
\begin{tabular}{@{}lccc@{}}
\toprule
\multirow{2}{*}{\textsc{\textbf{Method}}} & \textsc{\textbf{Medical}} & \textsc{\textbf{Agriculture}} & \textsc{\textbf{Remote Sensing}}  \\ \cmidrule(l){2-4} 
                                 & $S_{\alpha}$  (Kvasir)                  & IoU  (Leaf)                         & IoU  (Road)                          \\ \midrule
Teacher                          & 87.1\textsubscript{$\pm$ 0.12} & 42.7\textsubscript{$\pm$ 0.32} & 6.9\textsubscript{$\pm$ 0.13} \\
Student                          & 94.4\textsubscript{$\pm$ 0.06} & 76.8\textsubscript{$\pm$ 0.56} & 60.9\textsubscript{$\pm$ 0.14} \\\midrule
PKT                              & 94.0\textsubscript{$\pm$ 0.25} & 74.8\textsubscript{$\pm$ 0.14} & 57.3\textsubscript{$\pm$ 0.07} \\
VID                              & 94.1\textsubscript{$\pm$ 0.47} & 77.2\textsubscript{$\pm$ 0.37} & 61.1\textsubscript{$\pm$ 0.38} \\
ReviewKD                         & 93.4\textsubscript{$\pm$ 0.10} & 72.7\textsubscript{$\pm$ 0.37} & 55.9\textsubscript{$\pm$ 0.50} \\\midrule
TinySAM                          & 89.4\textsubscript{$\pm$ 0.10} & 45.2\textsubscript{$\pm$ 0.76} & 23.9\textsubscript{$\pm$ 2.61} \\
MobileSAM                        & 93.3\textsubscript{$\pm$ 0.15} & 74.1\textsubscript{$\pm$ 0.35} & 52.3\textsubscript{$\pm$ 0.46} \\
\textbf{InfoSAM2(Ours)}          & \textbf{94.5\textsubscript{$\pm$ 0.17}} & \textbf{77.3\textsubscript{$\pm$ 0.14}} & \textbf{61.3\textsubscript{$\pm$ 0.05}} \\
\bottomrule
\end{tabular}
\end{minipage}
\end{tabular*}
\end{scriptsize}
\end{center}
\end{table*}

    
    
    

\section{Experiments}
\label{experiments}

\textbf{Settings.} We conduct experiments using SAM~\cite{kirillov2023segment} (with a ViT-B backbone) and SAM2~\cite{ravi2024sam} (with a Hiera-B+ backbone) with Adapter~\cite{chen2022adaptformer,song2024simada}, and LoRA~\cite{hu2022lora} across four real-world domains: medical imaging, natural images, agriculture, and remote sensing. We fine-tune SAM's image encoder by adding adapters or LoRA, while fully training the decoder directly.
We use a batch size of 4 and the Adam optimizer with an initial learning rate of $2\times10^{-4}$, utilizing a CosineAnnealing scheduler that decays to a final learning rate of $2\times10^{-5}$. All the methods are trained for 10 epochs with structure loss (i.e., the combination of weighted IoU loss and binary cross entropy loss) unless otherwise specified. During training, prompts are randomly selected from noised ground truth boxes and points at a 1:1 ratio. During evaluation, ground truth boxes are used as the default geometric input prompts to ensure a fair comparison and minimize randomness. More implementation details are provided in Appendix~\ref{app_impl_detail}.

\textbf{Datasets.} In the natural image domain, we focus on camouflaged object segmentation~\cite{skurowski2018animal, le2019anabranch, fan2020camouflaged}. For medical imaging, we investigate polyp segmentation~\cite{bernal2015wm, jha2020kvasir} and skin lesion segmentation~\cite{codella2018skin}. In agriculture and remote sensing, we use leaf disease segmentation~\cite{rath2023leaf} and road segmentation datasets~\cite{mnih2013machine} as representative examples, respectively. For further details on the tasks and datasets, please refer to Appendix~\ref{app_data}.

To verify the effectiveness of our approach, we compare it with two categories of methods: PEFT methods and distillation methods. 

\textbf{PEFT Baselines.} The PEFT baselines encompass three types of methods: the direct application of SAM, PEFT methods from the NLP or CV domain, and PEFT methods designed for SAM. These are as follows: 1)~The zero-shot performance of the original SAM. 2)~Fine-tune SAM’s mask decoder only.  3)~BitFit~\cite{ben-zaken-etal-2022-bitfit}, which only fine-tunes bias terms in the pre-trained model. 4)~AdaptFormer~\cite{chen2022adaptformer}, which inserts the trainable bottleneck layers into the MLP block of the transformer. 5)~LoRA~\cite{hu2022lora} inserts trainable bottleneck layers parallel to the frozen linear weight. 6)~HQSAM~\cite{ke2024segment}, which introduces a learnable high-quality output token and enhances mask details by fusing mask decoder features with both early and final ViT features. 7)~SU-SAM~\cite{song2024simada} presents a simple framework that efficiently fine-tunes the SAM using Adapter or LoRA. 8)~ConvLoRA-SAM~\cite{zhong2024convolution} injects image-related inductive biases into the image encoder of SAM by integrating ultra-lightweight convolutional parameters into LoRA.

\textbf{Distillation Baselines.}  In this study, we compare our method with the following baselines: 1)~Logit-based distillation~\cite{zhu2018knowledge}. 2)~single-layer paired feature distillation (i.e., PKD~\cite{cao2022pkd}, PKT~\cite{passalis2020probabilistic}), which uses one-stage feature to distill knowledge, with MobileSAM~\cite{zhang2023faster} belonging to this category. 3)~multiple-layers paired feature distillation (i.e., VID~\cite{ahn2019variational}, IBD~\cite{kuang2023improving}), which utilizes multi-stage information to transfer knowledge, with each layer aligned separately. Similarly, TinySAM~\cite{shu2025tinysam} employs full-stage distillation. 4)~cross-layer feature distillation (i.e., SemCKD~\cite{chen2021cross}, ReviewKD~\cite{chen2021distilling}), which utilizes knowledge from multiple layers of the teacher model to supervise the student, by leveraging diverse information extracted from these layers. Currently, no work in SAM has explored cross-layer fusion distillation for PEFT. InfoSAM is the first to address this.

We report the main experimental results on representative datasets from different domains. Additional experimental results are provided in Appendix~\ref{app:additional_results}, and visualization results are available in Appendix~\ref{app_vis_results}. All experiments are conducted three times to mitigate randomness, with both average values and standard errors reported.

\subsection{Segment Anything Across Diverse Domains}

We compare InfoSAM with two categories of methods: PEFT methods and distillation methods. The results are presented in Table~\ref{tab:peft_comparison} and Table~\ref{tab:distillation_comparison}, respectively. 
In Table~\ref{tab:peft_comparison}, all PEFT methods outperform both the zero-shot performance and decoder-only fine-tuning, highlighting the importance of unified fine-tuning for SAM. Additionally, InfoSAM outperforms other PEFT techniques across various datasets from different domains. Compared to other PEFT methods, InfoSAM preserves the pre-trained, domain-invariant knowledge through information-based distillation, which proves effective in enhancing segmentation performance. 

In Table~\ref{tab:distillation_comparison}, it is noteworthy that most distillation methods are detrimental during PEFT, leading to worse performance compared to fine-tuning without distillation. Specifically, TinySAM employs full-stage distillation, requiring the student's features to fully mimic the teacher at every stage. However, this becomes catastrophic when the teacher performs poorly (e.g., achieving only 7.2\% IoU on the Road dataset). In contrast, InfoSAM further enhances PEFT performance in this challenging scenario, likely due to the relation compression process during distillation, which ensures the student model learns only the essential information from the teacher model.

\subsection{Extended Experiment with SAM2}
Note that our method is orthogonal to model development, making it easily transferable to SAM2 backbones. As shown in Table~\ref{tab:peft_and_distill_comparison_sam2}, InfoSAM demonstrates consistent effectiveness with SAM2. This transferability is attributed to InfoSAM's foundation in information-theoretic derivation, which is structure-independent.



\begin{figure}[t]
\begin{center}
    \centering
    
    \subfigure[\textcolor{black}{Performance on the Leaf dataset.}]{
    \includegraphics[width=0.45\textwidth]{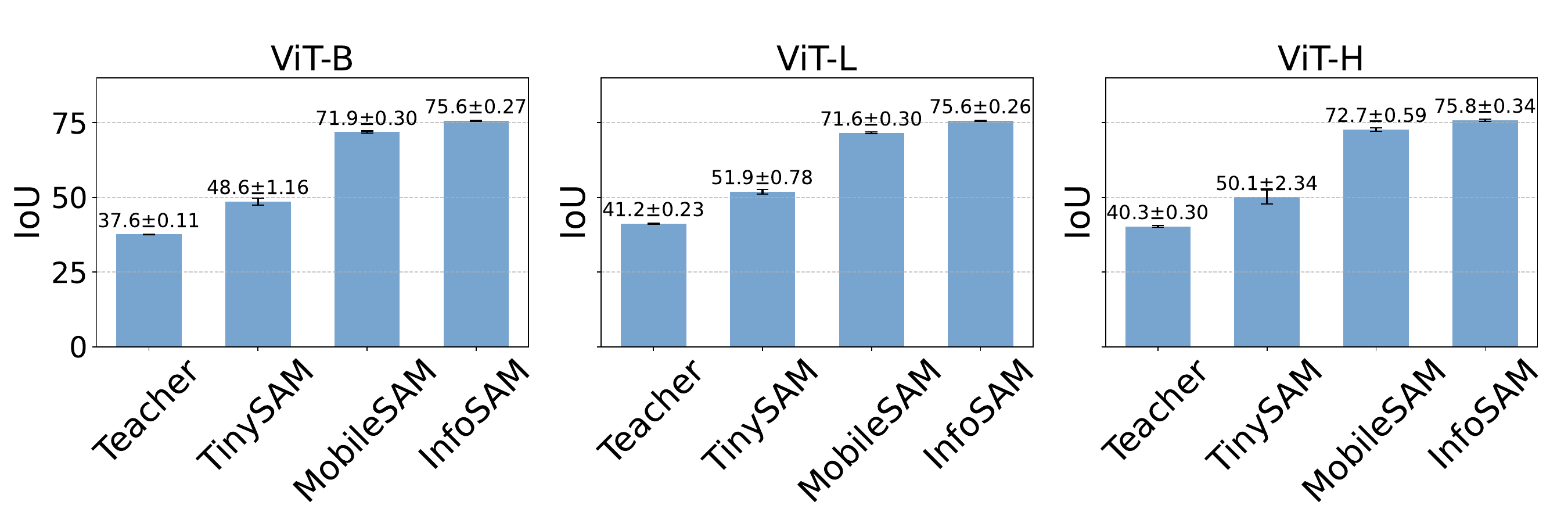}}
    
    \subfigure[\textcolor{black}{Performance on the Road dataset.}]{
    \includegraphics[width=0.45\textwidth]{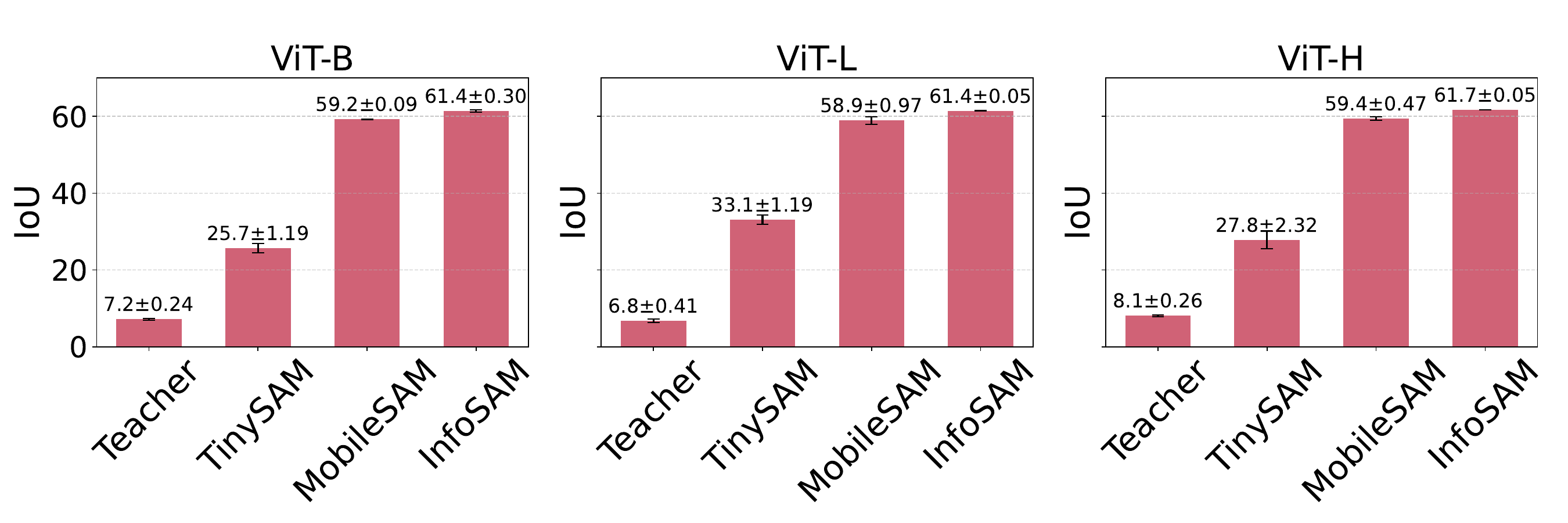}}
    
    \caption{\textbf{Performance of InfoSAM with larger teacher models} (i.e., ViT-L, ViT-H), while the student remains based on a ViT-B backbone. Each subfigure shows IoU metrics for different distillation methods on a specific dataset.}
    \label{fig:scale_up_tea}
\end{center}
\vskip -0.2in
\end{figure}

\subsection{Distillation Across Models of Different Sizes}

\begin{table}[t]
\caption{\textbf{Ablation study results of two losses}: relation compression loss $L_r$ and distillation loss $L_d$}
\label{tab:ablation_loss}
\begin{center}
\begin{scriptsize}
\setlength{\tabcolsep}{4pt}
\begin{tabular}{@{}p{0.8cm}p{0.8cm}ccc@{}}
\toprule
\multirow{2}{*}{$L_r$} & \multirow{2}{*}{$L_d$} & \textsc{\textbf{Medical}} & \textsc{\textbf{Agriculture}} & \textsc{\textbf{Remote Sensing}} \\ \cmidrule(l){3-5} 
                       &                        & $S_{\alpha}$  (Kvasir)   & IoU  (Leaf)                  & IoU  (Road)                    \\ \midrule
~ & ~   & 93.4  & 74.4 &  60.5 \\
~ & \checkmark  & 93.6 \textcolor{green}{(+0.2)} & 75.2 \textcolor{green}{(+0.8)} & 61.0 \textcolor{green}{(+0.5)} \\
\checkmark & \checkmark & 94.4 \textcolor{green}{(+1.0)} & 75.6 \textcolor{green}{(+1.2)} & 61.4 \textcolor{green}{(+0.9)} \\
\bottomrule
\end{tabular}
\end{scriptsize}
\end{center}
\end{table}

\begin{table}[t]
\caption{\textbf{Effects of the Relation Module (RM).} Enhancing TinySAM and MobileSAM using RM.}
\begin{center}
\begin{scriptsize}
\setlength{\tabcolsep}{6pt}
\begin{tabular}{@{}l|l|c|c@{}}
\toprule
\multirow{2}{*}{\textsc{\textbf{Model}}} 
& \multirow{2}{*}{\textsc{\textbf{Method}}} 
& \textsc{\textbf{Agriculture}} 
& \textsc{\textbf{Remote Sensing}} \\ 
\cmidrule(l){3-4}
                                 &                                  & IoU (Leaf)                   & IoU (Road)                    \\ \midrule
\multirow{2}{*}{TinySAM}         & w/o RM                          & 48.6\textsubscript{$\pm$ 1.14} & 28.7\textsubscript{$\pm$ 1.69} \\
                                 & w RM                            & \textbf{50.3\textsubscript{$\pm$ 0.76}} & \textbf{33.9\textsubscript{$\pm$ 0.32}} \\ \midrule

\multirow{2}{*}{MobileSAM}       & w/o RM                          & 71.9\textsubscript{$\pm$ 0.30} & 59.2\textsubscript{$\pm$ 0.09} \\
                                 & w RM                            & \textbf{73.8\textsubscript{$\pm$ 0.22}} & \textbf{61.3\textsubscript{$\pm$ 0.35}} \\ 
\bottomrule
\end{tabular}
\end{scriptsize}
\end{center}
\label{tab:effect_rm}
\vskip -0.1in
\end{table}

\begin{table}[t]
\caption{\textbf{Transferability of the Relation Module (RM).} InfoSAM-T represents the model trained with a pre-trained relation module from a different domain.}
\label{tab:transfer_rm}
\begin{center}
\begin{scriptsize}
\setlength{\tabcolsep}{4pt}
\begin{tabular}{@{}llcc@{}}
\toprule
\multirow{2}{*}{\textsc{\textbf{Method}}} & \textsc{\textbf{Frozen}}  & \textsc{\textbf{Medical}} & \textsc{\textbf{Agriculture}} \\ \cmidrule(l){3-4}
                                          & \textsc{\textbf{RM from}} & $S_{\alpha}$ (Kvasir)     & IoU (Leaf)                   \\ \midrule
InfoSAM                          & -                                         & 94.4\textsubscript{$\pm$ 0.12}                          & -                              \\
InfoSAM-T                        & Leaf                                      & 93.7\textsubscript{$\pm$ 0.24}                          & -                             \\
InfoSAM                          & -                                         &  -                        & 75.6\textsubscript{$\pm$ 0.30}                             \\
InfoSAM-T                        & Kvasir                                    &  -                        & 75.4\textsubscript{$\pm$ 0.45}                             \\
\bottomrule
\end{tabular}
\end{scriptsize}
\end{center}
\vskip -0.2in
\end{table}

We also verify the effectiveness of InfoSAM when scaling the teacher model to larger sizes. As shown in Fig.~\ref{fig:scale_up_tea}, InfoSAM shows comparable improvements to distillation methods specifically designed for SAM compression. It indicates that InfoSAM is better suited for PFET, even in traditional large-to-small knowledge distillation scenarios.

\subsection{Ablation Study}
\label{sec:abla_study}
\textbf{Ablation of Main Components.} We conduct an ablation study to evaluate the impact of InfoSAM's components: relation compression loss $\mathcal{L}_r$ and relation distillation loss $\mathcal{L}_d$ on three datasets: Kvasir, Leaf, and Road. Using SAM with an adapter as the baseline (Table~\ref{tab:ablation_loss}, row 1), row 2 introduces a fixed-dot relation module with mutual information-based distillation loss. Results show that incorporating simple relations improves performance, e.g., 0.8\% on Leaf, and mutual information effectively transfers relational features. Combining both losses yields further gains, e.g., 1.4\% on Leaf.

\textbf{Effects of Relation Module (RM).} 
To investigate the impact of the proposed relation module, we first conduct experiments to verify its effectiveness in enhancing various distillation methods.  Specifically, we evaluate and compare the performance of two compact models, MobileSAM and TinySAM, with and without integrating the relation module, as shown in Table~\ref{tab:effect_rm}.  We can observe a significant improvement with RM, e.g., 1.9\% IoU on the Leaf dataset.  These results suggest that the relation module can effectively capture and leverage high-level semantic information, thereby providing complementary benefits to existing distillation strategies during the fine-tuning stage.

Furthermore, Table~\ref{tab:transfer_rm} illustrates the effectiveness of domain-invariance dependencies in the relation module, we directly apply the module trained on one specific domain to another domain with entirely different knowledge. The results show that it still maintains satisfying results. \textcolor{black}{Moreover, we conduct experiments to explore the nature of domain-invariant information. We use the Boundary F1 Score~\cite{zhang2023learning} to evaluate such universal patterns. The results show that our methods employing the relation module perform better in preserving structural edge features. More results and analysis are available in Appendix \ref{app:deep_dive_rm}.}

\section{Conclusion}
\label{conclusion}
We introduce InfoSAM, an information-theoretic tuning framework designed for SAM adaptation. From an information bottleneck perspective, we extract domain-invariant knowledge from the pre-trained SAM and inject it into the fine-tuned SAM to enhance adaptation efficiency. Specifically, we first propose an attention-based module to capture structural relations while minimizing mutual information to retain the most essential ones. These relations are then transferred by maximizing their mutual information. Extensive evaluations across eight segmentation datasets spanning diverse domains and tasks strongly validate the effectiveness of InfoSAM.

\section*{Acknowledgments}
\label{acknowledgments}
This work was supported in part by the National Natural Science Foundation of China under Grants 62192781, 62172326, and 62137002; in part by the Project of China Knowledge Centre for Engineering Science and Technology; and supported by the Fundamental Research Funds for the Central Universities No. xxj032025002.

\section*{Impact Statement}
\label{impact}
This paper presents work whose goal is to advance the field of Machine Learning. There are many potential societal consequences of our work, none of which we feel must be specifically highlighted here.

\bibliography{ref}
\bibliographystyle{icml2025}

\newpage
\appendix
\onecolumn
\section{Derivation of information-theoretic Losses}
\label{app_ib}

\subsection{Multivariate Entropy and Mutual Information}
Following Definition 2 and Eq.(\ref{eq_joint_entropy}), we consider the matrix-based Rényi's $\alpha$-order joint entropy for multiple variables ~\cite{yu2019multivariate}.

\textbf{Definition 3.} Given a collection of \(n\) samples \(\{s_i = (x_1^i, x_2^i, \cdots, x_k^i)\}_{i=1}^n\), 
where the superscript \(i\) denotes the sample index, each sample contains \(k\) (\(k \geq 2\)) measurements 
\(x_1 \in \mathcal{X}_1, x_2 \in \mathcal{X}_2, \cdots, x_k \in \mathcal{X}_k\) obtained from the same realization, 
and the positive definite kernels \(\kappa_1 : \mathcal{X}_1 \times \mathcal{X}_1 \mapsto \mathbb{R}\), 
\(\kappa_2 : \mathcal{X}_2 \times \mathcal{X}_2 \mapsto \mathbb{R}\), \(\cdots\), 
\(\kappa_k : \mathcal{X}_k \times \mathcal{X}_k \mapsto \mathbb{R}\), 
a matrix-based analogue to Rényi’s \(\alpha\)-order joint-entropy among \(k\) variables can be defined as:
\begin{equation}
\begin{aligned}
\label{eq_app_multi_entropy}
S_\alpha(\mathbf{A}_1, \mathbf{A}_2, \cdots, \mathbf{A}_k) = 
S_\alpha\left(
\frac{\mathbf{A}_1 \circ \mathbf{A}_2 \circ \cdots \circ \mathbf{A}_k}
{\operatorname{tr}(\mathbf{A}_1 \circ \mathbf{A}_2 \circ \cdots \circ \mathbf{A}_k)}
\right)
\end{aligned}
\end{equation}
where \((\mathbf{A}_1)_{ij} = \kappa_1(x_1^i, x_1^j), (\mathbf{A}_2)_{ij} = \kappa_2(x_2^i, x_2^j), \cdots, (\mathbf{A}_k)_{ij} = \kappa_k(x_k^i, x_k^j)\), and \(\circ\) denotes the Hadamard product.

Following Definition 2 and Definition 3, mutual information can be extended to measure interactions among multiple variables by grouping them into sets and treating each set as a single variable. It can be defined as:
\begin{equation}
\begin{aligned}
\label{eq_app_multi_mi}
\mathbf{I}_\alpha\left(\mathbf{A}_1, \cdots, \mathbf{A}_k ; \mathbf{B}\right)=\mathbf{S}_\alpha\left(\mathbf{A}_1, \cdots, \mathbf{A}_k \right)+\mathbf{S}_\alpha(\mathbf{B})  -\mathbf{S}_\alpha\left(\mathbf{A}_1, \cdots, \mathbf{A}_k, \mathbf{B}\right)
\end{aligned}
\end{equation}
where $\mathbf{A}_1, \cdots, \mathbf{A}_k$ and $\mathbf{B}$ denote the normalized Gram matrices. 

\subsection{Derivation of Relation Compression Loss $\mathcal{L}_r$ and Relation Distillation Loss $\mathcal{L}_d$}
\label{app_derive_loss}

\textbf{Relation Compression Loss $\mathcal{L}_r$.}
In this paper, according to Eq.(\ref{eq_app_multi_entropy}) and Eq.(\ref{eq_app_multi_mi}), the mutual information of the compression process ($k=2$) can be defined as:
\begin{equation}
\begin{aligned}
\mathbf{I}_\alpha\left(\mathbf{A}_1, \mathbf{A}_2 ; \mathbf{B}\right)= \mathbf{S}_\alpha\left(\mathbf{A}_1, \mathbf{A}_2 \right)+\mathbf{S}_\alpha(\mathbf{B})  -\mathbf{S}_\alpha\left(\mathbf{A}_1, \mathbf{A}_2, \mathbf{B}\right)
\end{aligned}
\end{equation}
In the relation compression loss $\mathcal{L}_r$, the matrices $\mathbf{A}_1$, $\mathbf{A}_2$, and $\mathbf{B}$ are normalized Gram matrices constructed from the image embeddings $z_{i}^T$, the mask embeddings $z_{m}^T$, and the relation module outputs $r^T$, respectively. To maintain consistency with the previously defined relation compression loss (refer to Eq.(\ref{eq_loss_r_mi})), we replace $\mathbf{A}_1$, $\mathbf{A}_2$, and $\mathbf{B}$ with $G^{T}_{i}$, $G^{T}_{m}$, and $G^{T}_r \in \mathbb{R}^{N \times N}$, respectively, where all Gram matrices are computed using the polynomial kernel function defined as:
\begin{equation}
\begin{aligned}
 \textcolor{black}{\kappa(x, y) = x^\top y,}
\end{aligned}
\end{equation}
where $\mathbf{x}, \mathbf{y} \in \mathbb{R}^N$ are input vectors. For instance, $G^{T}_{i}$ is computed as the normalized Gram matrix:
\begin{equation}
\begin{aligned}
 \textcolor{black}{G^{T}_{i} = \frac{\kappa(z^T_i, z^T_i)}{\text{tr}(\kappa(z^T_i, z^T_i))},}
\end{aligned}
\end{equation}
where $\text{tr}(\cdot)$ denotes the trace of the matrix.

Furthermore, Rényi’s $\alpha$-order entropy is reformulated using its eigenvalue expansion (refer to Eq.(\ref{eq_eigs})), leading to:
\begin{equation}
\begin{aligned}
\label{app_eq_loss_r_mi}
\mathcal{L}_r =& \mathbf{I}_\alpha(z^{T}_{i}, z^{T}_{m};r^{T}) \\
=&\cancel{\mathbf{S}_\alpha(G^{T}_{i}, G^{T}_{m})}+ \mathbf{S}_\alpha(G^{T}_r)-\mathbf{S}_\alpha(G^{T}_{i}, G^{T}_{m},G^{T}_r) \\
=& 
 \textcolor{black}{\frac{1}{1 - \alpha} \log_2 \sum_{i=1}^{n} \lambda_i^\alpha\left(G_{r}^T\right)
-
\frac{1}{1 - \alpha} \log_2 \sum_{i=1}^{n} \lambda_i^\alpha\left(G_{imr}^T\right)}
\end{aligned}
\end{equation}
where $G_{imr}^T$ is also normalized to have a trace of one. The teacher entropy term is excluded from this loss because the teacher's weights remain fixed throughout the training process. Substituting the marginal and joint entropy definitions from Definition 1 and Definition 3, $G^{S}_{imr}=G^{T}_{i} \circ G^{T}_{m} \circ G^{T}_{r}$. The $\circ$ is Hadamard product. 

Since computing the eigenvalues of large matrices is typically computationally expensive during training~\cite{kerr2009qr}, we restrict the value of $\alpha$ to 2. This choice allows us to use the Frobenius norm as a proxy objective function. Notably, the Frobenius norm has a connection with the eigenspectrum. Specifically, for a symmetric matrix $\mathbf{A}$, its Frobenius norm can be expressed as:
\begin{equation}
\begin{aligned}
\label{eq_app_eigs_equal}
 \textcolor{black}{\|\mathbf{A}\|_F^2 = \mathrm{tr}(\mathbf{A}\mathbf{A}^H) = \sum_{i=1}^n \lambda_i(\mathbf{A}^2)=\sum_{i=1}^n \lambda_i^2(\mathbf{A})}
\end{aligned}
\end{equation}
where $\mathrm{tr}(\cdot)$ denotes the trace operation. Since $\mathbf{A}$ is a symmetric matrix, $\sum_{i=1}^n \lambda_i(\mathbf{A}^2)$ is equivalent to $\sum_{i=1}^n \lambda_i^2(\mathbf{A})$~\cite{dong2023optimal}. Through this formulation, the Frobenius norm not only simplifies the computation but also retains an intrinsic connection to the matrix eigenspectrum, which is significant for both model training and theoretical analysis. Given $G_{r}^T$ and $G_{imr}^T$ is positive semi-definite, $\mathcal{L}_r$ can be reformulated as: 
\begin{equation}
\begin{aligned}
\mathcal{L}_r = & 
 \textcolor{black}{\frac{1}{1 - 2} \log_2 \sum_{i=1}^{n} \lambda_i^2 \left(G_{r}^T\right)
-
\frac{1}{1 - 2} \log_2 \sum_{i=1}^{n} \lambda_i^2 \left(G_{imr}^T\right)} \\
=&- \log_2 \| G^{T}_{r} \|_F^2 + \log_2 \| G^{T}_{imr} \|_F^2
\end{aligned}
\end{equation}
The two terms in $\mathcal{L}_{r}$ serve distinct purposes.  First, the term $- \log_2 \| G^T_r \|_F^2$ imposes spectral compression on the relation module. This ensures that the relation module focuses on the most discriminative features while suppressing irrelevant variations. Second, the term $\log_2 \| G^T_{imr} \|_F^2$ achieves joint entropy minimization across the image encoder, mask decoder, and relation module. It filters out spurious relationships induced by domain-specific noise while preserving domain-invariant interactions encoded in $G^T_{imr}$. This is critical for cross-domain adaptation, as it maintains consistency in feature interactions across different data distributions.

\textbf{Relation Distillation Loss $\mathcal{L}_d$.}
Similar to $\mathcal{L}_r$, the goal of distillation is to maximize the mutual information of teacher-student relations. The mutual information of the distillation process can be defined as:
\begin{equation}
\begin{aligned}
\mathbf{I}_\alpha\left(r^{T};r^{S}\right)= &\mathbf{S}_\alpha\left(G^{T}_r \right)+\mathbf{S}_\alpha(G^{S}_r)  -\mathbf{S}_\alpha\left(G^{T}_{r}, G^{S}_{r}\right) \\
= & 
 \textcolor{black}{\frac{1}{1 - \alpha} \log_2 \sum_{i=1}^{n} \lambda_i^\alpha\left(G_{r}^T\right) +
\frac{1}{1 - \alpha} \log_2 \sum_{i=1}^{n} \lambda_i^\alpha\left(G_{r}^S\right)
-
\frac{1}{1 - \alpha} \log_2 \sum_{i=1}^{n} \lambda_i^\alpha\left(G_{r}^{TS}\right)}
\end{aligned}
\end{equation}
where $G_{r}^S$, $G_{r}^T$ and $G_{r}^{TS}=G_{r}^T \circ G_{r}^S$ is the normalized Gram matrices. Then, applying Eq.(\ref{eq_app_eigs_equal}) and set $\alpha=2$, the distillation loss is:
\begin{equation}
\begin{aligned}
\mathcal{L}_d =& -\mathbf{I}_\alpha\left(r^{T};r^{S}\right) \\
= & \textcolor{black}{-\frac{1}{1 - \alpha} \log_2 \sum_{i=1}^{n} \lambda_i^\alpha\left(G_{r}^T\right)
- \frac{1}{1 - \alpha} \log_2 \sum_{i=1}^{n} \lambda_i^\alpha\left(G_{r}^S\right)
+ \frac{1}{1 - \alpha} \log_2 \sum_{i=1}^{n} \lambda_i^\alpha\left(G_{r}^{TS}\right)} \\
=&  \log_2 \| G^{T}_{r} \|_F^2 + \log_2 \| G^{S}_{r} \|_F^2 - \log_2 \| G^{TS}_{r} \|_F^2
\end{aligned}
\end{equation}
Overall, the mutual information loss balances feature constraints by regularizing the student's relational feature complexity to prevent overfitting while aligning the student's and teacher's relational distributions for effective knowledge transfer. The three terms in the distillation loss serve distinct purposes: $\log_2 \| G^{T}_{r} \|_F^2$ captures the complexity of the teacher's relational features, guiding the teacher to learn rich and detailed representations rather than overly simplified ones; $\log_2 \| G^{S}_{r} \|_F^2 $ regularizes the student's feature complexity, preventing overfitting by keeping the representations manageable; $- \log_2 \| G^{TS}_{r} \|_F^2$ Aligns the relational distributions between the teacher and student, ensuring the student learns the relationships between features as captured by the teacher.

\definecolor{commentcolor}{rgb}{0.3451, 0.6706, 0.3451} 
\definecolor{keywordcolor}{rgb}{0.9608, 0.3961, 0.7294}   
\definecolor{stringcolor}{rgb}{0.58,0,0.82}  

\lstset{
  language=Python,
  basicstyle=\ttfamily\footnotesize,
  keywordstyle=\color{keywordcolor}\bfseries,
  commentstyle=\color{commentcolor},
  stringstyle=\color{stringcolor},
  numbers=none, 
  numberstyle=\tiny\color{commentcolor},
  stepnumber=1,
  numbersep=8pt,
  frame=none, 
  backgroundcolor=\color{white},
  showstringspaces=false,
  tabsize=2,
  captionpos=b,
  breaklines=true,
  breakatwhitespace=false,
  escapeinside={(*@}{@*)}
}

\subsection{Pseudocode of InfoSAM}
\label{pseudocode}
Here, we summarize the core fine-tuning process for SAM using the proposed information-theoretic loss in Algorithm.\ref{algo1}. Relationships between features across different network components are modeled and normalized to capture meaningful interactions. These interactions drive two key loss functions: a compression loss that strengthens the student network's ability to represent complex features and a distillation loss that aligns the student's relational understanding with the teacher's. This method emphasizes relational and structural learning to achieve better generalization and accuracy.

\begin{algorithm}[H]
\caption{PyTorch-style pseudocode for InfoSAM}
\label{algo1}
\begin{lstlisting}[language=Python]
# F_t, F_s: Pre-trained SAM (teacher) and fine-tuned SAM (student)
# z_t_i, z_s_i: The output of the teacher and student image encoders
# z_t_m, z_s_m: The output tokens in the mask decoder of the teacher and student 
# f_t, f_s: Teacher and student relation modules
# y_t, y_s: Teacher and student outputs
# y: Ground-truth labels
# Frob: Function for computing the square of the Frobenius norm

for x, y in loader:
    # Forward pass
    z_t_i, z_t_m, y_t = F_t(x)
    z_s_i, z_s_m, y_s = F_s(x)

    # Compute structure loss
    loss_ce = struct_loss(y_s,y)

    # Compute relations between image encoder and mask decoder
    f_s = f_t
    r_t = f_t(z_t_i, z_t_m)
    r_s = f_s(z_s_i, z_s_m)

    # Normalize the representations
    z_t_i_norm = F.normalize(z_t_i, p=2)
    z_t_m_norm = F.normalize(z_t_m, p=2)
    
    # Compute normalized Gram matrices for compression loss_r
    G_t_i = matmul(z_t_i_norm, z_t_i_norm.T)
    G_t_m = matmul(z_t_m_norm, z_t_m_norm.T)
    G_t_r = matmul(r_t, r_t.T)
    G_t_r_norm = G_t_r / trace(G_t_r)
    G_t_imr_norm = G_t_i * G_t_m * G_t_r / trace(G_t_i * G_t_m * G_t_r) 

    # Compute normalized Gram matrices for distillation loss_d
    G_s_r = matmul(r_s, r_s.T)
    G_s_r_norm = G_s_r / trace(G_s_r)
    G_ts_r_norm = G_s_r * G_t_r / trace(G_s_r * G_t_r)

    # Compute relation compression loss_r and distillation loss_d
    loss_r = - log2(Frob(G_t_r_norm)) + log2(Frob(G_t_imr_norm))
    loss_d = log2(Frob(G_t_r_norm)) + log2(Frob(G_s_r_norm)) - log2(Frob(G_ts_r_norm))
    loss_info = lamda_1 * loss_r + lamda_2 * loss_d

    # The overall loss
    loss = loss_ce + loss_info

    # Optimization step
    loss.backward()
    optimizer.step()
\end{lstlisting}
\end{algorithm}

\section{Implementation Details}
\label{app_impl_detail}

\subsection{Architectures of Segment Anything Model (SAM)}
SAM~\cite{kirillov2023segment} consists of three key modules, i.e., image encoder, prompt encoder, and mask decoder. The image encoder is a heavy ViT-based network for image feature extraction. The prompt encoder is designed to capture positional information from geometric prompts (i.e., points, boxes, or masks) to generate prompt embeddings. The mask decoder, a two-way transformer module, combines image embeddings and prompt tokens to generate the final mask. The released model, trained on 11 million images and 1 billion high-quality masks, demonstrates impressive zero-shot capability in handling various conventional natural images. The latest SAM2~\cite{ravi2024sam} introduces a significant evolution over its predecessor by extending its capabilities to the domain of video segmentation. SAM2 replaces the backbone of SAM with Hiera backbones. And introduces an additional streaming memory component designed for processing video frames. However, recent studies\cite{wu2025medical,ji2024segment} have revealed that SAM performs poorly in real-world segmentation tasks across domains such as medicine, agriculture, and remote sensing. Fine-tuning SAM for downstream tasks has been widely recommended.

\subsection{Architectures of Adapter and LoRA}

\begin{figure*}[t]
\vskip 0.2in
\begin{center}
    \centering
    \subfigure[\textcolor{black}{Adapter}]{
    \includegraphics[width=0.4\textwidth]{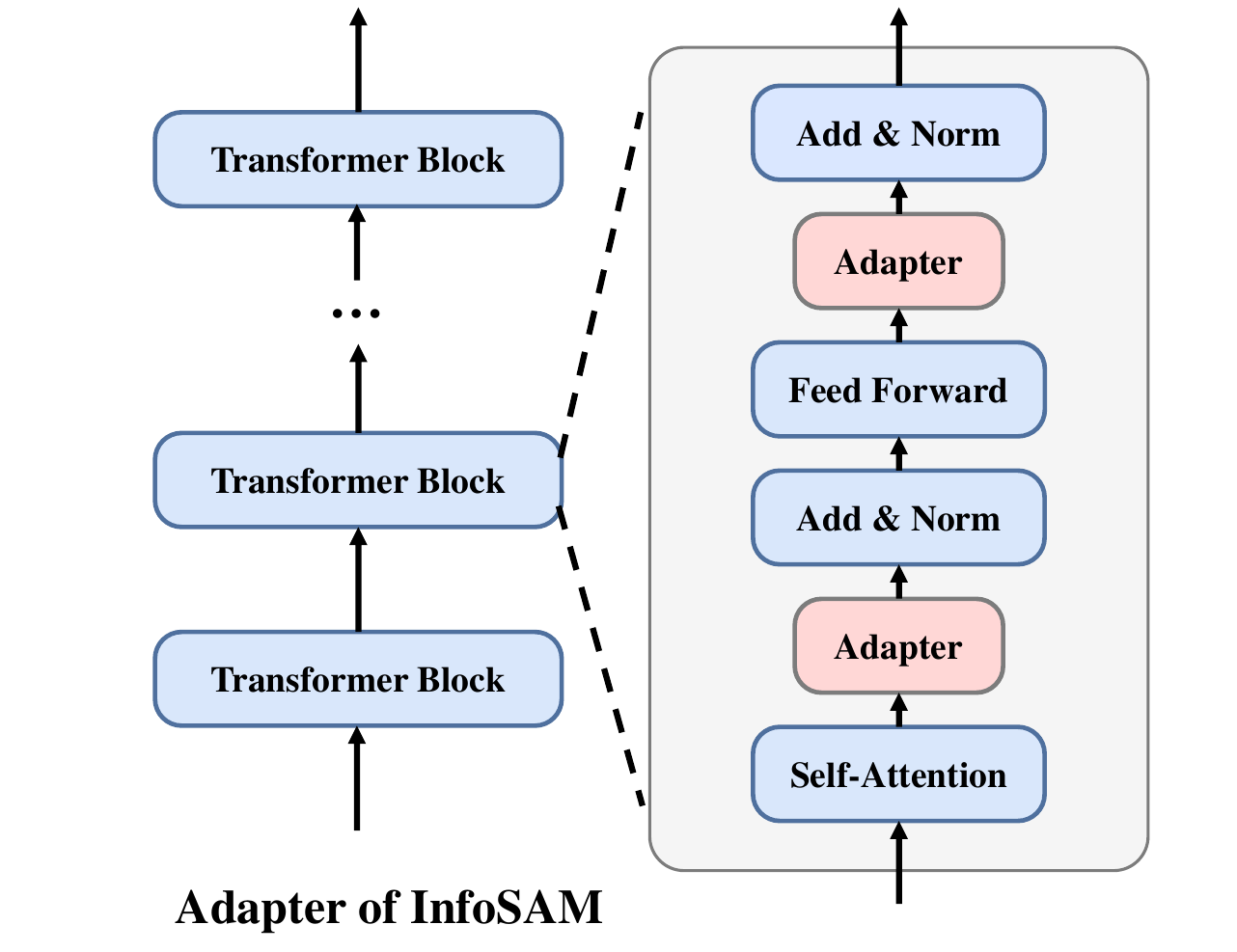}
    \label{app_fig_adapter}}
    \subfigure[\textcolor{black}{LoRA}]{
    \includegraphics[width=0.4\textwidth]{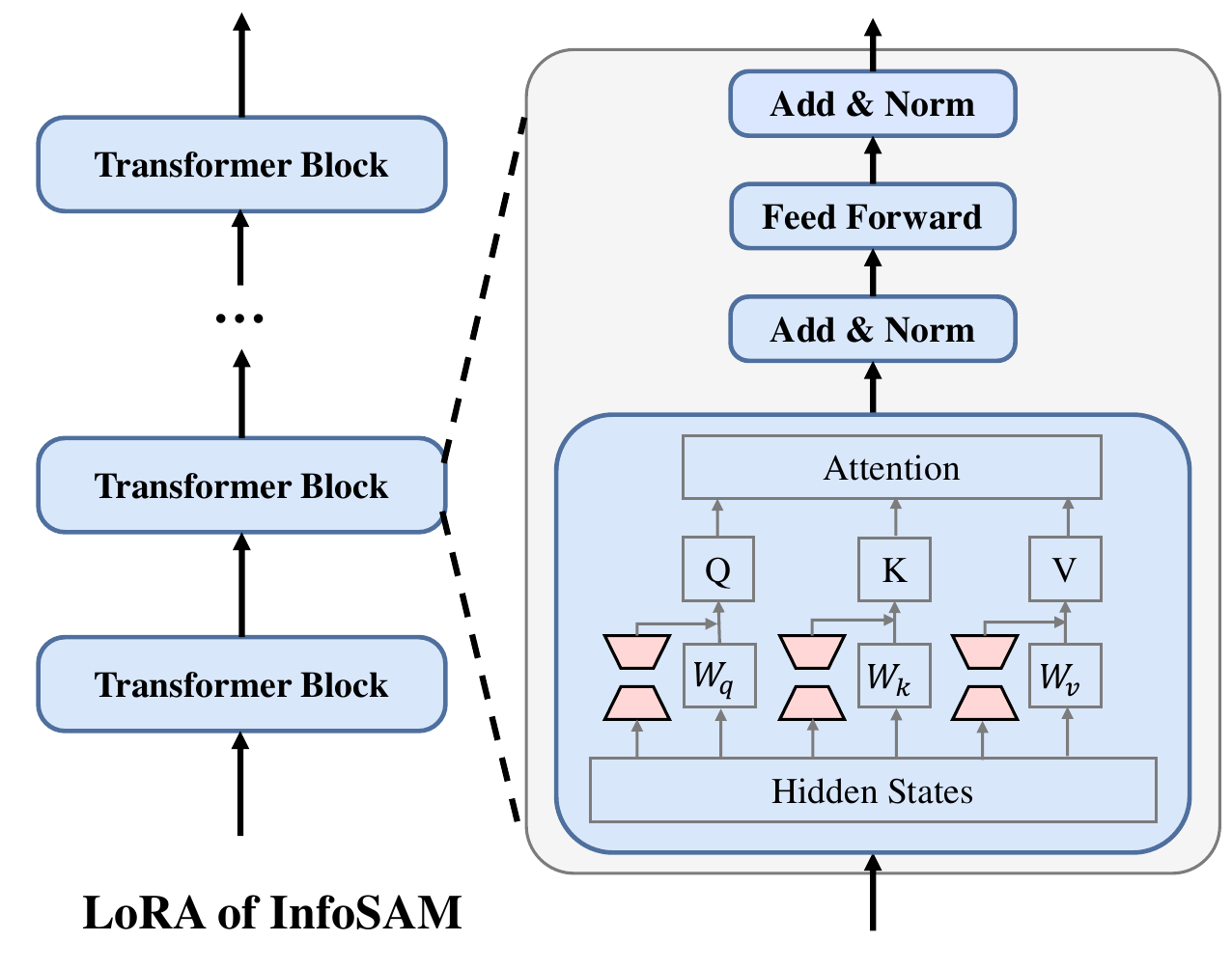}
    \label{app_fig_lora}}
    \caption{The architecture of InfoSAM with Adapter and LoRA}
\end{center}
\vskip -0.2in
\end{figure*}

We implement InfoSAM with two of the most widely adopted PEFT methods: Adapter~\cite{chen2022adaptformer,song2024simada} and LoRA~\cite{hu2022lora}. We fine-tune the image encoder of SAM and SAM2 by incorporating adapters or LoRA, while fully training the mask decoder. Notably, all training hyperparameters used for SAM and SAM2 remain the same.

The Adapter method introduces lightweight modules into the Transformer architecture by adding small projection layers between the original layers. Specifically, the adapter module consists of two linear transformations: a down-projection and an up-projection, with a non-linear activation layer (GELU in our implementation) applied between them. Let the input feature be $\mathbf{X} \in \mathbb{R}^{B \times L \times D}$, where $B$ is the batch size, $L$ is the sequence length, and $D$ is the feature dimension. The adapter performs the following operations:
\begin{equation}
\mathbf{z} = \mathbf{X} \mathbf{W}_\text{down}
\end{equation}
\begin{equation}
\mathbf{z}_\text{act} = \text{GELU}(\mathbf{z})
\end{equation}
\begin{equation}
\mathbf{h} = \mathbf{z}_\text{act} \mathbf{W}_\text{up}
\end{equation}
where $\mathbf{W}_\text{down} \in \mathbb{R}^{D \times r}$ and $\mathbf{W}_\text{up} \in \mathbb{R}^{r \times D}$ are learnable weight matrices, and $r$ denotes the bottleneck dimension, typically a small fraction of $D$. In our experiments, we set the bottleneck ratio $r/D$ to 0.25. The final output of the adapter is computed as:
\begin{equation}
\mathbf{X}_\text{out} = \mathbf{X} + \mathbf{h}
\end{equation}
with a residual connection preserving the original information.
As illustrated in Fig.~\ref{app_fig_adapter}, two adapter modules are sequentially inserted between the attention and FFN layers.

LoRA (Low-Rank Adaptation) is another parameter-efficient method that integrates lightweight modules into the Transformer architecture. Similar to the adapter method, LoRA introduces trainable down-projection and up-projection matrices. However, as shown in Fig.~\ref{app_fig_lora}, instead of being applied between layers, LoRA is added directly to the query ($\mathbf{q}$), key ($\mathbf{k}$), and value ($\mathbf{v}$) projections within the self-attention mechanism.

Given the input feature $\mathbf{X} \in \mathbb{R}^{B \times L \times D}$, the multi-head self-attention mechanism with $H$ heads computes the query, key, and value as follows: 
\begin{equation} 
\mathbf{Q}_h = \mathbf{X} \mathbf{W}_{Q,h}, \quad \mathbf{K}_h = \mathbf{X} \mathbf{W}_{K,h}, \quad \mathbf{V}_h = \mathbf{X} \mathbf{W}_{V,h}, \quad h = 1, \dots, H 
\end{equation} 
where $\mathbf{W}_{Q,h}, \mathbf{W}_{K,h}, \mathbf{W}_{V,h} \in \mathbb{R}^{D \times D_h}$, and $D_h = D / H$ is the dimensionality of each attention head.

LoRA introduces low-rank updates to each head's query, key, and value projections. Specifically: \begin{equation} \mathbf{Q}_h' = \mathbf{Q}_h + \Delta \mathbf{Q}_h, \quad \mathbf{K}_h' = \mathbf{K}_h + \Delta \mathbf{K}_h, \quad \mathbf{V}_h' = \mathbf{V}_h + \Delta \mathbf{V}_h \end{equation} where the low-rank updates are defined as: 
\begin{equation} 
\Delta \mathbf{Q}_h = \mathbf{X} \mathbf{W}_{Q,h,\text{down}} \mathbf{W}_{Q,h,\text{up}}, \quad \Delta \mathbf{K}_h = \mathbf{X} \mathbf{W}_{K,h,\text{down}} \mathbf{W}_{K,h,\text{up}}, \quad \Delta \mathbf{V}_h = \mathbf{X} \mathbf{W}_{V,h,\text{down}} \mathbf{W}_{V,h,\text{up}} 
\end{equation} 
Here:
$\mathbf{W}_{Q,h,\text{down}}, \mathbf{W}_{K,h,\text{down}}, \mathbf{W}_{V,h,\text{down}} \in \mathbb{R}^{D \times r}$,
$\mathbf{W}_{Q,h,\text{up}}, \mathbf{W}_{K,h,\text{up}}, \mathbf{W}_{V,h,\text{up}} \in \mathbb{R}^{r \times D_h}$.
Note that the low-rank projection matrices are head-specific, and $r$ is the bottleneck dimension shared across all heads, determining the rank of the adaptation. In our experiments, we set $r = 4$ for all the LoRA-based methods.

\section{Tasks and Datasets}
\label{app_data}
We benchmark InfoSAM across eight diverse benchmarks spanning four different domains, following Conv-LoRA~~\cite{zhong2024convolution}. The specific segmentation tasks and corresponding datasets utilized for benchmarking are elaborated below.

\subsection{Natural Image}

\textbf{Camouflaged Object Segmentation.} The task aims to detect objects hidden within complex or visually cluttered backgrounds, posing greater challenges compared to traditional object segmentation. We use three camouflaged object detection datasets: COD10K~\cite{fan2020camouflaged}, CHAMELEON~\cite{skurowski2018animal}, and CAMO~\cite{le2019anabranch}. COD10K contains 3,040 training and 2,026 testing samples, CHAMELEON provides 76 testing images, and CAMO includes 1,000 training and 250 testing images. The combined dataset of COD10K and CAMO training images is used, with 10\% randomly split for validation, and testing is performed on all three datasets.

\subsection{Medical Image}
\textbf{Polyp Segmentation.} The task of polyp segmentation in gastrointestinal endoscopic images is critical for early colorectal cancer diagnosis and treatment planning, posing significant challenges due to the considerable variability in polyp shapes and sizes. We selecte two polyp segmentation datasets: Kvasir~\cite{jha2020kvasir} and CVC-ClinicDB (also known as CVC-612)~\cite{bernal2015wm}. The Kvasir dataset consists of 1,000 images, while CVC-ClinicDB contains 612 publicly accessible images. \cite{fan2020pranet} splits the images into a 9:1 ratio for training and testing. Additionally, 20\% of the training set is randomly selected as a validation set for use during training.

\textbf{Skin Lesion Segmentation.} The task involves identifying different types of skin lesions in medical images, playing a vital role in the early diagnosis and treatment of skin conditions, particularly skin cancer. However, it remains challenging due to ambiguous boundaries and color variations. We select the ISIC 2017 dataset~\cite{codella2018skin} for skin lesion segmentation, which contains 2,000 images for training, 150 images for validation, and 600 images for testing.

\subsection{Agriculture}
\textbf{Leaf Segmentation.} The task focuses on identifying individual plant leaves in agricultural images, supporting automation in plant disease control and high-quality food production. We use the Leaf Disease Segmentation dataset~\cite{rath2023leaf}, which includes 498 images for training and 90 for testing, with 20\% of the training set randomly split for validation.

\subsection{Remote Sensing}
\textbf{Road Segmentation.} This task involves detecting road regions in images or video frames, which is essential for autonomous driving, traffic analysis, and urban planning. We use the Massachusetts Roads Dataset~\cite{mnih2013machine}, containing 1,107 images for training, 13 for validation, and 48 for testing. All the methods are trained for 20 epochs. During validation and testing, we use 5-point prompts instead of noisy ground-truth box prompts, as the complexity of road structures in remote sensing images is challenging to capture with box-based prompts.

\section{Additional Experiment Results}
\label{app:additional_results}
Due to space constraints, experimental results that could not be included in the main body are provided here. These include additional results on more polyp segmentation and camouflaged object segmentation datasets using SAM, as well as the complete experimental results conducted on SAM2.

\subsection{Additional Results with SAM}
\textbf{Polyp Segmentation:} As shown in Table~\ref{app_tab:polyp}, InfoSAM exhibits superior performance in polyp segmentation on the Kvasir and CVC-612 datasets, outperforming both PEFT and distillation-based methods.
\begin{table}[H]
\caption{Additional results of polyp segmentation.}
\label{app_tab:polyp}
\begin{center}
\begin{scriptsize}
\setlength{\tabcolsep}{4pt}
\begin{tabular}{l|ccc|ccc}
\toprule
\multirow{2}{*}{\textsc{\textbf{Method}}} & \multicolumn{3}{c|}{\textbf{Kvasir}}         & \multicolumn{3}{c}{\textbf{CVC-612}}          \\ 
                        & $S_\alpha$ & $E_\phi$ & $F_\beta^w$ & $S_\alpha$ & $E_\phi$ & $F_\beta^w$ \\
\midrule
SAM                     & 83.0\textsubscript{$\pm$ 0.10} & 88.8\textsubscript{$\pm$ 0.29} & 79.7\textsubscript{$\pm$ 0.06} & 87.8\textsubscript{$\pm$ 0.19} & 94.5\textsubscript{$\pm$ 0.19} & 85.8\textsubscript{$\pm$ 0.34} \\
decoder-only            & 90.9\textsubscript{$\pm$ 0.05} & 95.2\textsubscript{$\pm$ 0.18} & 89.1\textsubscript{$\pm$ 0.48} & 92.9\textsubscript{$\pm$ 0.14} & 97.4\textsubscript{$\pm$ 0.26} & 89.5\textsubscript{$\pm$ 0.63} \\
BitFit                  & 92.5\textsubscript{$\pm$ 0.12} & 96.3\textsubscript{$\pm$ 0.20} & 91.0\textsubscript{$\pm$ 0.37} & 94.0\textsubscript{$\pm$ 0.39} & 98.5\textsubscript{$\pm$ 0.21} & 91.8\textsubscript{$\pm$ 0.81} \\
AdaptFormer             & 93.3\textsubscript{$\pm$ 0.68} & 97.0\textsubscript{$\pm$ 0.81} & 92.8\textsubscript{$\pm$ 0.94} & 95.2\textsubscript{$\pm$ 0.15} & 99.0\textsubscript{$\pm$ 0.14} & 93.8\textsubscript{$\pm$ 0.45} \\
LoRA                    & 93.0\textsubscript{$\pm$ 0.14} & 96.6\textsubscript{$\pm$ 0.11} & 91.8\textsubscript{$\pm$ 0.57} & 94.3\textsubscript{$\pm$ 0.31} & 98.7\textsubscript{$\pm$ 0.17} & 92.3\textsubscript{$\pm$ 0.62} \\
Adapter                 & 93.4\textsubscript{$\pm$ 0.12} & 97.1\textsubscript{$\pm$ 0.15} & 92.9\textsubscript{$\pm$ 0.13} &  95.1\textsubscript{$\pm$ 0.51} & 98.8\textsubscript{$\pm$ 0.40} & 94.2\textsubscript{$\pm$ 0.33} \\
HQ-SAM                  & 91.1\textsubscript{$\pm$ 0.50} & 95.5\textsubscript{$\pm$ 0.57} & 89.9\textsubscript{$\pm$ 0.68} & 93.2\textsubscript{$\pm$ 0.51} & 98.0\textsubscript{$\pm$ 0.60} & 90.9\textsubscript{$\pm$ 1.01} \\
SU-SAM                  & 93.8\textsubscript{$\pm$ 0.02} & 97.5\textsubscript{$\pm$ 0.06} & \textbf{94.1\textsubscript{$\pm$ 0.45}} & 95.3\textsubscript{$\pm$ 0.42} & 98.9\textsubscript{$\pm$ 0.39} & 94.4\textsubscript{$\pm$ 0.65} \\
ConvLoRA-SAM            & 92.9\textsubscript{$\pm$ 0.13} & 96.6\textsubscript{$\pm$ 0.28} & 92.2\textsubscript{$\pm$ 0.46} & 94.4\textsubscript{$\pm$ 0.05} & 98.8\textsubscript{$\pm$ 0.14} & 92.6\textsubscript{$\pm$ 0.63} \\
\midrule
Logit                   & 93.2\textsubscript{$\pm$ 0.19} & 96.5\textsubscript{$\pm$ 0.19} & 92.9\textsubscript{$\pm$ 0.11} & 95.1\textsubscript{$\pm$ 0.30} & 99.0\textsubscript{$\pm$ 0.34} & 93.8\textsubscript{$\pm$ 0.59} \\
PKD                     & 92.2\textsubscript{$\pm$ 0.25} & 96.0\textsubscript{$\pm$ 0.17} & 91.9\textsubscript{$\pm$ 0.17} & 94.2\textsubscript{$\pm$ 0.25} & 98.5\textsubscript{$\pm$ 0.13} & 92.3\textsubscript{$\pm$ 0.40} \\
PKT                     & 93.7\textsubscript{$\pm$ 0.41} & 97.3\textsubscript{$\pm$ 0.53} & 93.8\textsubscript{$\pm$ 0.43} & 95.3\textsubscript{$\pm$ 0.19} & 99.1\textsubscript{$\pm$ 0.18} & 94.3\textsubscript{$\pm$ 0.36} \\
IBD                     & 91.5\textsubscript{$\pm$ 0.14} & 95.3\textsubscript{$\pm$ 0.05} & 89.9\textsubscript{$\pm$ 0.74} & 93.1\textsubscript{$\pm$ 0.13} & 97.4\textsubscript{$\pm$ 0.15} & 90.0\textsubscript{$\pm$ 0.52} \\
VID                     & 93.7\textsubscript{$\pm$ 0.16} & 97.4\textsubscript{$\pm$ 0.07} & 93.4\textsubscript{$\pm$ 0.49} & 95.0\textsubscript{$\pm$ 0.36} & 98.7\textsubscript{$\pm$ 0.15} & 93.8\textsubscript{$\pm$ 0.20} \\
SemCKD                  & 92.4\textsubscript{$\pm$ 0.07} & 96.2\textsubscript{$\pm$ 0.03} & 91.2\textsubscript{$\pm$ 0.52} & 93.9\textsubscript{$\pm$ 0.55} & 98.3\textsubscript{$\pm$ 0.44} & 91.5\textsubscript{$\pm$ 0.54} \\
ReviewKD                & 92.4\textsubscript{$\pm$ 0.33} & 96.4\textsubscript{$\pm$ 0.26} & 91.6\textsubscript{$\pm$ 0.65} & 94.0\textsubscript{$\pm$ 0.59} & 98.4\textsubscript{$\pm$ 0.49} & 92.2\textsubscript{$\pm$ 1.40} \\
MobileSAM               & 92.5\textsubscript{$\pm$ 0.12} & 96.3\textsubscript{$\pm$ 0.14} & 91.4\textsubscript{$\pm$ 0.15} & 94.6\textsubscript{$\pm$ 0.27} & 98.6\textsubscript{$\pm$ 0.12} & 92.4\textsubscript{$\pm$ 0.77} \\
TinySAM                 & 88.5\textsubscript{$\pm$ 0.31} & 93.5\textsubscript{$\pm$ 0.24} & 86.0\textsubscript{$\pm$ 0.79} & 92.1\textsubscript{$\pm$ 0.42} & 96.4\textsubscript{$\pm$ 0.59} & 88.1\textsubscript{$\pm$ 0.79} \\
\textbf{InfoSAM(Ours)}  & \textbf{94.4\textsubscript{$\pm$ 0.12}} & \textbf{97.9\textsubscript{$\pm$ 0.09}} & 93.9\textsubscript{$\pm$ 0.09} & \textbf{95.3\textsubscript{$\pm$ 0.09}} & \textbf{98.9\textsubscript{$\pm$ 0.09}} &\textbf{ 94.3\textsubscript{$\pm$ 0.15}} \\
\bottomrule
\end{tabular}
\end{scriptsize}
\end{center}
\end{table}

\textbf{Camouflaged Object Segmentation:} As illustrated in Table~\ref{app_tab:camouflaged}, InfoSAM achieves state-of-the-art performance in camouflaged object segmentation across the CHAMELEON, CAMO, and COD10K datasets, surpassing both PEFT and distillation-based approaches.
\begin{table}[H]
\caption{Additional results of camouflaged object segmentation.}
\label{app_tab:camouflaged}
\begin{center}
\begin{scriptsize}
\setlength{\tabcolsep}{4pt}
\begin{tabular}{l|ccc|ccc|ccc}
\toprule
\multirow{2}{*}{\textsc{\textbf{Method}}} & \multicolumn{3}{c|}{\textbf{CHAMELEON}}      & \multicolumn{3}{c|}{\textbf{CAMO}}          & \multicolumn{3}{c}{\textbf{COD10K}} \\ 
                        & $S_\alpha$ & $E_\phi$ & $F_\beta^w$ & $S_\alpha$ & $E_\phi$ & $F_\beta^w$ & $S_\alpha$ & $E_\phi$ & $F_\beta^w$ \\
\midrule
SAM                     & 80.4\textsubscript{$\pm$ 0.14} & 88.9\textsubscript{$\pm$ 0.10} & 74.5\textsubscript{$\pm$ 0.13} & 79.7\textsubscript{$\pm$ 0.02} & 88.8\textsubscript{$\pm$ 0.09} & 79.6\textsubscript{$\pm$ 0.01} & 83.5\textsubscript{$\pm$ 0.02} & 92.5\textsubscript{$\pm$ 0.03} & 79.2\textsubscript{$\pm$ 0.01}  \\
decoder-only            & 87.0\textsubscript{$\pm$ 0.20} & 93.5\textsubscript{$\pm$ 0.41} & 80.0\textsubscript{$\pm$ 0.74} & 84.9\textsubscript{$\pm$ 0.38} & 92.7\textsubscript{$\pm$ 0.34} & 81.8\textsubscript{$\pm$ 0.33} & 87.1\textsubscript{$\pm$ 0.10} & 94.4\textsubscript{$\pm$ 0.11} & 79.9\textsubscript{$\pm$ 0.13} \\
BitFit                  & 89.6\textsubscript{$\pm$ 0.47} & 96.1\textsubscript{$\pm$ 0.33} & 84.2\textsubscript{$\pm$ 0.77} & 87.5\textsubscript{$\pm$ 0.13} & 94.5\textsubscript{$\pm$ 0.08} & 85.3\textsubscript{$\pm$ 0.48} & 89.2\textsubscript{$\pm$ 0.31} & 95.9\textsubscript{$\pm$ 0.21} & 83.6\textsubscript{$\pm$ 0.59} \\
AdaptFormer             & 92.2\textsubscript{$\pm$ 0.13} & 97.6\textsubscript{$\pm$ 0.30} & 89.4\textsubscript{$\pm$ 0.55} & 87.9\textsubscript{$\pm$ 0.10} & 94.8\textsubscript{$\pm$ 0.21} & 86.2\textsubscript{$\pm$ 0.19} & 90.1\textsubscript{$\pm$ 0.16} & 96.5\textsubscript{$\pm$ 0.10} & 85.8\textsubscript{$\pm$ 0.55} \\
LoRA                    & 90.4\textsubscript{$\pm$ 0.48} & 96.5\textsubscript{$\pm$ 0.25} & 85.5\textsubscript{$\pm$ 0.59} & 87.7\textsubscript{$\pm$ 0.59} & 94.6\textsubscript{$\pm$ 0.50} & 85.1\textsubscript{$\pm$ 0.64} & 89.8\textsubscript{$\pm$ 0.17} & 96.3\textsubscript{$\pm$ 0.06} & 84.9\textsubscript{$\pm$ 0.40} \\
Adapter                 & 92.5\textsubscript{$\pm$ 0.10} & 97.9\textsubscript{$\pm$ 0.11} & 90.1\textsubscript{$\pm$ 0.39} & 88.2\textsubscript{$\pm$ 0.44} & 94.8\textsubscript{$\pm$ 0.34} & 86.7\textsubscript{$\pm$ 0.92} & 90.2\textsubscript{$\pm$ 0.25} & 96.5\textsubscript{$\pm$ 0.19} & 86.0\textsubscript{$\pm$ 0.58} \\
HQ-SAM                  & 87.0\textsubscript{$\pm$ 0.34} & 93.3\textsubscript{$\pm$ 0.54} & 79.7\textsubscript{$\pm$ 0.07} & 85.1\textsubscript{$\pm$ 0.10} & 92.6\textsubscript{$\pm$ 0.10} & 81.0\textsubscript{$\pm$ 0.61} & 87.3\textsubscript{$\pm$ 0.14} & 94.5\textsubscript{$\pm$ 0.25} & 80.0\textsubscript{$\pm$ 0.49} \\
SU-SAM                  & 92.3\textsubscript{$\pm$ 0.28} & 97.9\textsubscript{$\pm$ 0.27} & \textbf{90.0\textsubscript{$\pm$ 0.23}} & 88.3\textsubscript{$\pm$ 0.21} & 95.0\textsubscript{$\pm$ 0.22} & 86.2\textsubscript{$\pm$ 0.59} & 90.2\textsubscript{$\pm$ 0.15} & 96.5\textsubscript{$\pm$ 0.09} & 86.0\textsubscript{$\pm$ 0.45} \\
ConvLoRA-SAM            & 90.6\textsubscript{$\pm$ 0.34} & 96.6\textsubscript{$\pm$ 0.19} & 86.0\textsubscript{$\pm$ 0.71} & 87.5\textsubscript{$\pm$ 0.39} & 94.5\textsubscript{$\pm$ 0.17} & 85.4\textsubscript{$\pm$ 0.41} & 89.8\textsubscript{$\pm$ 0.23} & 96.2\textsubscript{$\pm$ 0.21} & 84.9\textsubscript{$\pm$ 0.61} \\
\midrule
Logit                   & 91.9\textsubscript{$\pm$ 0.47} & 97.4\textsubscript{$\pm$ 0.46} & 89.2\textsubscript{$\pm$ 0.55} & 88.4\textsubscript{$\pm$ 0.08} & 94.9\textsubscript{$\pm$ 0.05} & 87.1\textsubscript{$\pm$ 0.22} & 90.4\textsubscript{$\pm$ 0.12} & 96.6\textsubscript{$\pm$ 0.10} & 86.1\textsubscript{$\pm$ 0.31} \\
PKD                     & 90.7\textsubscript{$\pm$ 0.40} & 96.2\textsubscript{$\pm$ 0.64} & 87.5\textsubscript{$\pm$ 1.10} & 87.0\textsubscript{$\pm$ 0.43} & 94.1\textsubscript{$\pm$ 0.23} & 84.3\textsubscript{$\pm$ 0.97} & 89.7\textsubscript{$\pm$ 0.17} & 96.3\textsubscript{$\pm$ 0.16} & 85.4\textsubscript{$\pm$ 0.43} \\
PKT                     & 92.2\textsubscript{$\pm$ 0.40} & 97.7\textsubscript{$\pm$ 0.50} & 90.0\textsubscript{$\pm$ 0.73} & 87.8\textsubscript{$\pm$ 0.40} & 94.5\textsubscript{$\pm$ 0.35} & 86.2\textsubscript{$\pm$ 0.46} & 90.3\textsubscript{$\pm$ 0.16} & 96.6\textsubscript{$\pm$ 0.10} & 86.2\textsubscript{$\pm$ 0.50} \\
IBD                     & 86.6\textsubscript{$\pm$ 0.29} & 93.0\textsubscript{$\pm$ 0.60} & 79.5\textsubscript{$\pm$ 0.77} & 85.2\textsubscript{$\pm$ 0.47} & 92.6\textsubscript{$\pm$ 0.35} & 82.4\textsubscript{$\pm$ 0.31} & 87.9\textsubscript{$\pm$ 0.13} & 94.8\textsubscript{$\pm$ 0.18} & 81.4\textsubscript{$\pm$ 0.37} \\
VID                     & 92.2\textsubscript{$\pm$ 0.18} & 97.9\textsubscript{$\pm$ 0.13} & 89.8\textsubscript{$\pm$ 0.40} & 87.9\textsubscript{$\pm$ 0.22} & 94.8\textsubscript{$\pm$ 0.34} & 86.3\textsubscript{$\pm$ 0.32} & 90.3\textsubscript{$\pm$ 0.16} & 96.5\textsubscript{$\pm$ 0.16} & 86.2\textsubscript{$\pm$ 0.10} \\
SemCKD                  & 88.8\textsubscript{$\pm$ 0.27} & 95.1\textsubscript{$\pm$ 0.29} & 83.4\textsubscript{$\pm$ 0.60} & 86.2\textsubscript{$\pm$ 0.16} & 93.5\textsubscript{$\pm$ 0.21} & 82.8\textsubscript{$\pm$ 1.54} & 88.5\textsubscript{$\pm$ 0.31} & 95.6\textsubscript{$\pm$ 0.18} & 83.1\textsubscript{$\pm$ 0.34} \\
ReviewKD                & 90.3\textsubscript{$\pm$ 0.30} & 96.5\textsubscript{$\pm$ 0.19} & 85.6\textsubscript{$\pm$ 0.02} & 86.7\textsubscript{$\pm$ 0.07} & 94.0\textsubscript{$\pm$ 0.09} & 84.6\textsubscript{$\pm$ 0.63} & 89.4\textsubscript{$\pm$ 0.23} & 96.0\textsubscript{$\pm$ 0.10} & 84.3\textsubscript{$\pm$ 0.42} \\
MobileSAM               & 91.4\textsubscript{$\pm$ 0.32} & 96.7\textsubscript{$\pm$ 0.21} & 88.6\textsubscript{$\pm$ 0.41} & 87.1\textsubscript{$\pm$ 0.36} & 94.1\textsubscript{$\pm$ 0.27} & 85.1\textsubscript{$\pm$ 0.09} & 90.1\textsubscript{$\pm$ 0.12} & 96.6\textsubscript{$\pm$ 0.08} & 86.1\textsubscript{$\pm$ 0.29} \\
TinySAM                 & 84.1\textsubscript{$\pm$ 0.09} & 90.3\textsubscript{$\pm$ 0.61} & 76.6\textsubscript{$\pm$ 0.30} & 83.7\textsubscript{$\pm$ 0.39} & 91.6\textsubscript{$\pm$ 0.31} & 81.1\textsubscript{$\pm$ 0.35} & 86.3\textsubscript{$\pm$ 0.39} & 94.0\textsubscript{$\pm$ 0.11} & 80.0\textsubscript{$\pm$ 0.47} \\
\textbf{InfoSAM(Ours)}  & \textbf{92.6\textsubscript{$\pm$ 0.14}} & \textbf{98.0\textsubscript{$\pm$ 0.25}} & 89.6\textsubscript{$\pm$ 0.22} & \textbf{88.5\textsubscript{$\pm$ 0.05}} & \textbf{95.1\textsubscript{$\pm$ 0.05}} & \textbf{87.1\textsubscript{$\pm$ 0.09}} & \textbf{90.5\textsubscript{$\pm$ 0.05}} & \textbf{96.7\textsubscript{$\pm$ 0.09}} & \textbf{86.2\textsubscript{$\pm$ 0.22}} \\ 
\bottomrule
\end{tabular}
\end{scriptsize}
\end{center}
\end{table}

\subsection{Additional Results with SAM2}
The complete experimental results of InfoSAM with the SAM2 backbone are provided in Table~\ref{app_tab:peft_comparison_sam2} and Table~\ref{app_tab:distill_comparison_sam2}. Maintaining the strong performance of SAM, InfoSAM achieves superior results with SAM2.
\begin{table}[H]
\caption{Complete comparison results of PEFT methods with SAM2 across different domains.}
\label{app_tab:peft_comparison_sam2}
\begin{center}
\begin{scriptsize}
\setlength{\tabcolsep}{4pt}
\begin{tabular}{@{}l|ccc|cc|cc@{}}
\toprule
\multirow{3}{*}{\textsc{\textbf{Method}}} & \multicolumn{3}{c|}{\textsc{\textbf{Medical}}} & \multicolumn{2}{c|}{\textsc{\textbf{Agriculture}}} & \multicolumn{2}{c}{\textsc{\textbf{Remote Sensing}}}                     \\ \cmidrule(l){2-8} 
                                 & \multicolumn{3}{c|}{Kvasir}                                                                      & \multicolumn{2}{c|}{Leaf}                                       & \multicolumn{2}{c}{Road}                                        \\
                                 & $S_{\alpha}$                   & $E_{\phi}$                     & $F_{\beta}^{\omega}$           & IoU                            & Dice                           & IoU                            & Dice                           \\ \midrule
SAM2                             & 87.1\textsubscript{$\pm$ 0.12}          & 90.2\textsubscript{$\pm$ 0.06}          & 85.2\textsubscript{$\pm$ 0.20}          & 42.7\textsubscript{$\pm$ 0.32}          & 53.3\textsubscript{$\pm$ 0.32}          & 6.9\textsubscript{$\pm$ 0.13}           & 12.4\textsubscript{$\pm$ 0.37}          \\
decoder-only                     & 93.2\textsubscript{$\pm$ 0.07}          & 96.6\textsubscript{$\pm$ 0.05}          & 92.1\textsubscript{$\pm$ 0.41}          & 71.8\textsubscript{$\pm$ 0.58}          & 82.2\textsubscript{$\pm$ 0.58}          & 48.5\textsubscript{$\pm$ 0.47}          & 64.7\textsubscript{$\pm$ 0.49}          \\ \midrule
BitFit                           & 93.8\textsubscript{$\pm$ 0.09}          & 97.0\textsubscript{$\pm$ 0.06}          & 93.2\textsubscript{$\pm$ 0.17}          & 75.4\textsubscript{$\pm$ 0.29}          & 85.2\textsubscript{$\pm$ 0.26}          & 59.2\textsubscript{$\pm$ 0.26}          & 74.0\textsubscript{$\pm$ 0.25}          \\
AdaptFormer                      & 93.7\textsubscript{$\pm$ 0.19}          & 97.2\textsubscript{$\pm$ 0.42}          & 93.3\textsubscript{$\pm$ 0.38}          & 73.6\textsubscript{$\pm$ 1.10}          & 83.7\textsubscript{$\pm$ 0.78}          & 59.9\textsubscript{$\pm$ 0.35}          & 74.6\textsubscript{$\pm$ 0.28}          \\
LoRA                             & 93.7\textsubscript{$\pm$ 0.10}          & 97.0\textsubscript{$\pm$ 0.07}          & 93.2\textsubscript{$\pm$ 0.37}          & 75.9\textsubscript{$\pm$ 0.40}          & 85.5\textsubscript{$\pm$ 0.36}          & 60.8\textsubscript{$\pm$ 0.32}          & 75.3\textsubscript{$\pm$ 0.29}          \\
Adapter                          & 94.4\textsubscript{$\pm$ 0.06}          & 97.5\textsubscript{$\pm$ 0.09}          & 93.8\textsubscript{$\pm$ 0.03}          & 76.8\textsubscript{$\pm$ 0.56}          & 86.2\textsubscript{$\pm$ 0.50}          & 60.9\textsubscript{$\pm$ 0.14}          & 75.4\textsubscript{$\pm$ 0.11}          \\ \midrule
\textbf{LoRA+Ours}               & \textbf{94.0\textsubscript{$\pm$ 0.09}} & \textbf{97.0\textsubscript{$\pm$ 0.08}} & \textbf{93.4\textsubscript{$\pm$ 0.25}} & \textbf{76.1\textsubscript{$\pm$ 0.38}} & \textbf{85.7\textsubscript{$\pm$ 0.31}} & \textbf{60.9\textsubscript{$\pm$ 0.05}} & \textbf{75.5\textsubscript{$\pm$ 0.08}} \\
\textbf{Adapter+Ours}            & \textbf{94.5\textsubscript{$\pm$ 0.17}} & \textbf{97.4\textsubscript{$\pm$ 0.16}} & \textbf{94.0\textsubscript{$\pm$ 0.16}} & \textbf{77.3\textsubscript{$\pm$ 0.14}} & \textbf{86.6\textsubscript{$\pm$ 0.08}} & \textbf{61.3\textsubscript{$\pm$ 0.05}} & \textbf{75.8\textsubscript{$\pm$ 0.05}} \\ \bottomrule
\end{tabular}
\end{scriptsize} 
\end{center}
\end{table}

\begin{table}[H]
\caption{Complete comparison results of distillation methods with SAM2 across various domains.}
\label{app_tab:distill_comparison_sam2}
\begin{center}
\begin{scriptsize}
\setlength{\tabcolsep}{4pt}
\begin{tabular}{@{}l|ccc|cc|cc@{}}
\toprule
\multirow{3}{*}{\textsc{\textbf{Method}}} & \multicolumn{3}{c|}{\textsc{\textbf{Medical}}} & \multicolumn{2}{c|}{\textsc{\textbf{Agriculture}}} & \multicolumn{2}{c}{\textsc{\textbf{Remote Sensing}}}                     \\ \cmidrule(l){2-8} 
                                 & \multicolumn{3}{c|}{Kvasir}                                                 & \multicolumn{2}{c|}{Leaf}                                       & \multicolumn{2}{c}{Road}                                        \\
                                 & $S_{\alpha}$      & $E_{\phi}$ & $F_\beta^w$         & IoU                            & Dice                           & IoU                            & Dice                           \\ \midrule
Teacher                          & 87.1\textsubscript{$\pm$ 0.12} & 90.2\textsubscript{$\pm$ 0.06} & 85.2\textsubscript{$\pm$ 0.20} & 42.7\textsubscript{$\pm$ 0.32} & 53.3\textsubscript{$\pm$ 0.32} & 6.9\textsubscript{$\pm$ 0.13}  & 12.4\textsubscript{$\pm$ 0.37} \\
Student                          & 94.4\textsubscript{$\pm$ 0.06} & 97.5\textsubscript{$\pm$ 0.09} & 93.8\textsubscript{$\pm$ 0.03} & 76.8\textsubscript{$\pm$ 0.56} & 86.2\textsubscript{$\pm$ 0.50} & 60.9\textsubscript{$\pm$ 0.14} & 75.4\textsubscript{$\pm$ 0.11} \\ \midrule
PKT                              & 94.0\textsubscript{$\pm$ 0.25} & 97.2\textsubscript{$\pm$ 0.10} & 93.7\textsubscript{$\pm$ 0.40} & 74.8\textsubscript{$\pm$ 0.14} & 84.7\textsubscript{$\pm$ 0.20} & 57.3\textsubscript{$\pm$ 0.07} & 72.5\textsubscript{$\pm$ 0.04} \\
VID                              & 94.1\textsubscript{$\pm$ 0.47} & 97.2\textsubscript{$\pm$ 0.45} & 93.5\textsubscript{$\pm$ 0.45} & 77.2\textsubscript{$\pm$ 0.37} & 86.4\textsubscript{$\pm$ 0.26} & 61.1\textsubscript{$\pm$ 0.38} & 75.6\textsubscript{$\pm$ 0.30} \\
ReviewKD                         & 93.4\textsubscript{$\pm$ 0.10} & 97.0\textsubscript{$\pm$ 0.10} & 92.7\textsubscript{$\pm$ 0.34} & 72.7\textsubscript{$\pm$ 0.37} & 83.0\textsubscript{$\pm$ 0.36} & 55.9\textsubscript{$\pm$ 0.50} & 71.3\textsubscript{$\pm$ 0.53} \\ \midrule
MobileSAM                        & 93.3\textsubscript{$\pm$ 0.15} & 96.7\textsubscript{$\pm$ 0.06} & 92.5\textsubscript{$\pm$ 0.69} & 74.1\textsubscript{$\pm$ 0.35} & 84.1\textsubscript{$\pm$ 0.18} & 52.3\textsubscript{$\pm$ 0.46} & 68.3\textsubscript{$\pm$ 0.38} \\
TinySAM                          & 89.4\textsubscript{$\pm$ 0.10} & 93.3\textsubscript{$\pm$ 0.32} & 86.3\textsubscript{$\pm$ 0.20} & 45.2\textsubscript{$\pm$ 0.76} & 56.1\textsubscript{$\pm$ 0.63} & 23.9\textsubscript{$\pm$ 2.61} & 36.5\textsubscript{$\pm$ 3.22} \\
\textbf{InfoSAM2(Ours)}          & \textbf{94.5\textsubscript{$\pm$ 0.17}} & \textbf{97.4\textsubscript{$\pm$ 0.16}} & \textbf{94.0\textsubscript{$\pm$ 0.16}} & \textbf{77.3\textsubscript{$\pm$ 0.14}} & \textbf{86.6\textsubscript{$\pm$ 0.08}} & \textbf{61.3\textsubscript{$\pm$ 0.05}} & \textbf{75.8\textsubscript{$\pm$ 0.05}} \\ \bottomrule
\end{tabular}
\end{scriptsize}
\end{center}
\end{table}

\section{Hyper-parameter Sensitivity Analysis}
\label{app_hyer}

\subsection{Analysis of the $\lambda_1$ and $\lambda_2$ in $\mathcal{L}_{info}$}
\begin{figure}[H]
\vskip 0.2in
\begin{center}
    \centering
    \subfigure[\textcolor{black}{Kvasir}]{
    \includegraphics[width=0.23\textwidth]{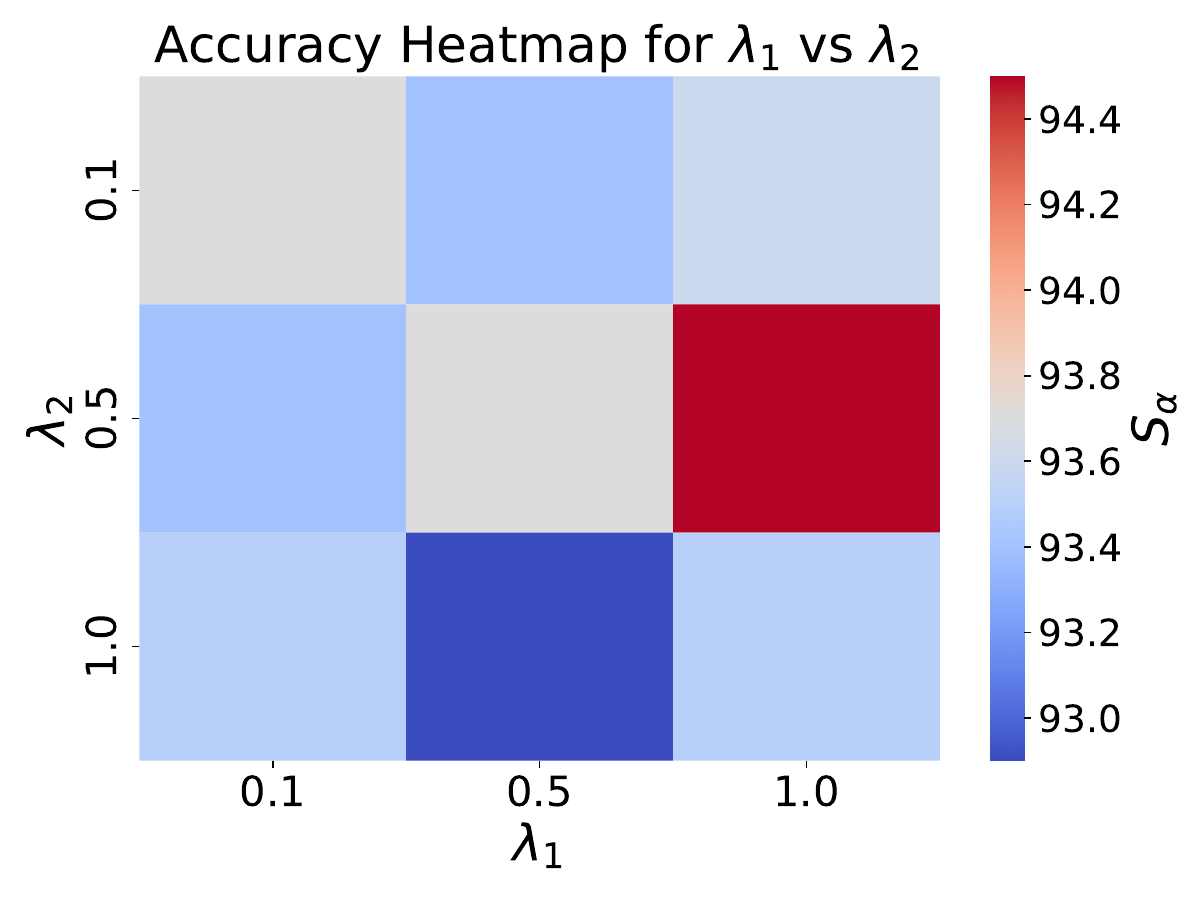}}
    \subfigure[\textcolor{black}{Leaf}]{
    \includegraphics[width=0.23\textwidth]{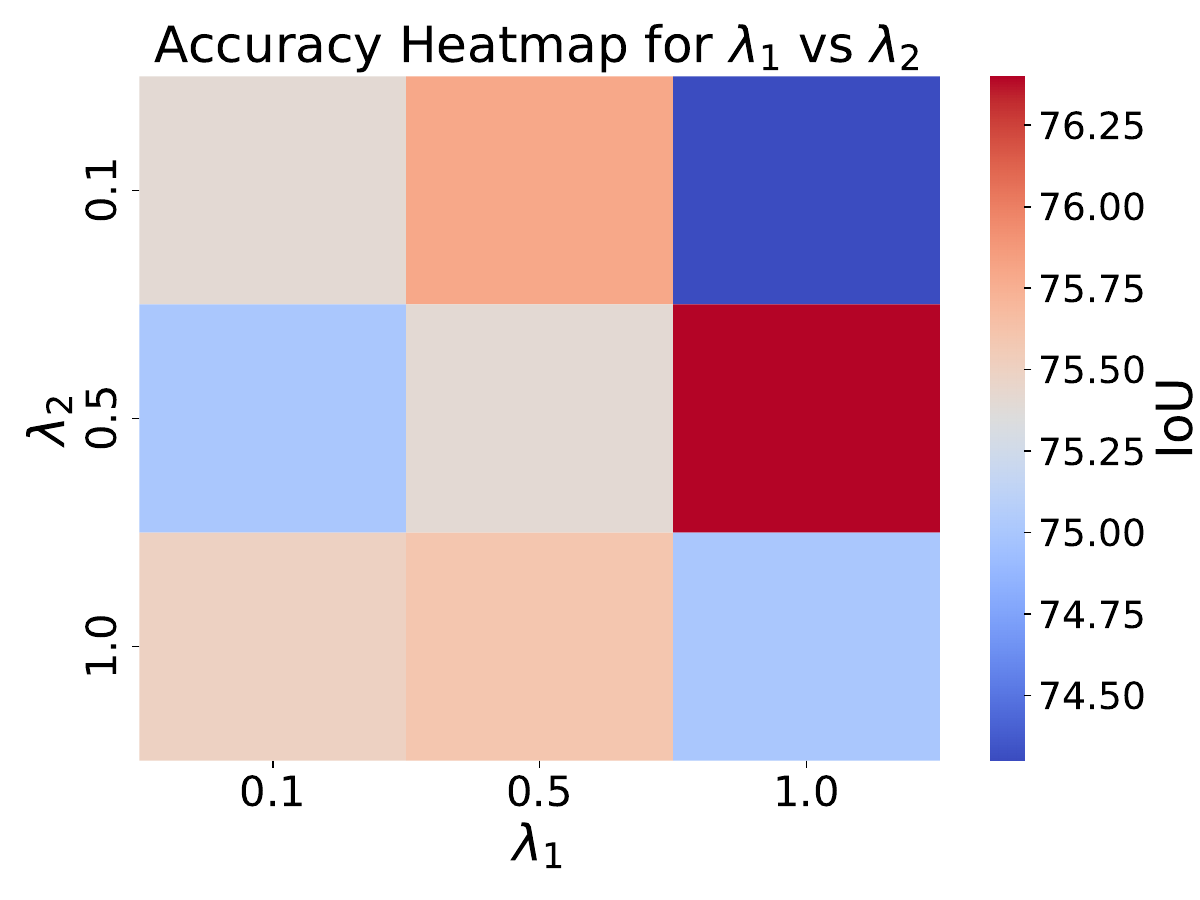}}
    \subfigure[\textcolor{black}{Road}]{
    \includegraphics[width=0.23\textwidth]{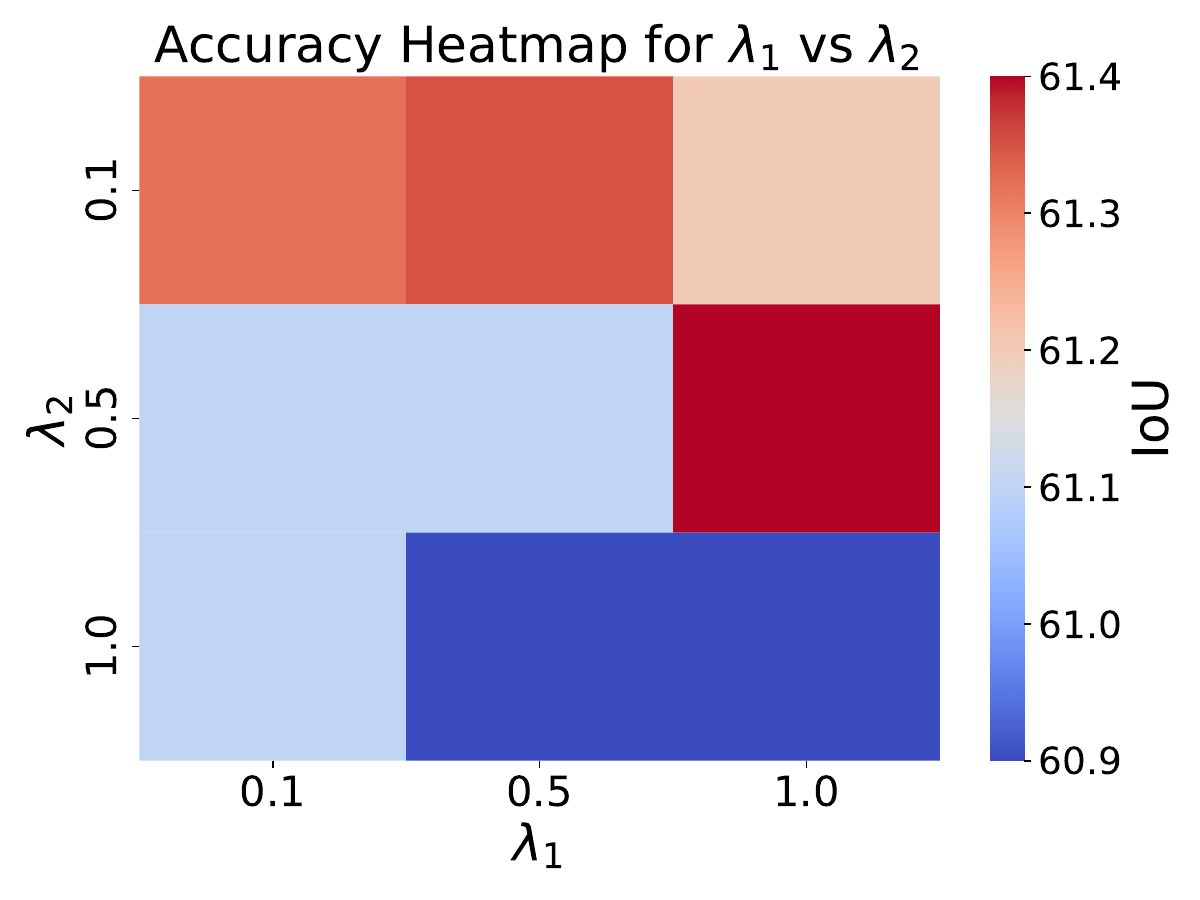}}
    \caption{Hyper-parameter sensitivity study of $\lambda_1$ and $\lambda_2$ in balancing $L_r$ and $L_d$, with Kvasir, Leaf, and Road datasets (Best viewed in color).}
    \label{fig_ablation_lambda}
\end{center}
\vskip -0.2in
\end{figure}

In Fig.~\ref{fig_ablation_lambda}, we conduct a key hyper-parameter sensitivity study of $\lambda_1$ and $\lambda_2$ in balancing relation compression loss $L_r$ and relation distillation loss $L_d$ across three typical domains. Each sub-figure shows the heat map under different hyper-parameter settings, reflecting the change of loss function. It is recommended to view it in color display for best results. According to the accuracy heat map, we set $\lambda_1=1, \lambda_2=0.5$. 

 \textcolor{black}{\subsection{Analysis of the $\alpha$ Parameter in Matrix-based Rényi's Entropy}}
 \textcolor{black}{In this paper, we set $\alpha=2$ to compute matrix-based Rényi's $\alpha$-entropy via the Frobenius norm. The core reasons for choosing $\alpha=2$ in matrix-based Rényi's $\alpha$-entropy are as follows: 
(i) The primary practical motivations are computational efficiency and alignment with prior works. By setting $\alpha=2$, we enable direct computation of matrix-based Rényi entropy through Frobenius norm operations (see Eq.(\ref{eq_lossr})), eliminating the necessity for eigenvalue decomposition. This optimization reduces time complexity from $O(n^3)$ to $O(n^2)$ ($n$ represents the sample numbers)~\cite{dong2023optimal}, substantially reducing computational costs while maintaining theoretical rigor, particularly advantageous for high-dimensional data analysis~\cite{yu2019multivariate}. Additionally, prior research has successfully applied Rényi entropy with $\alpha=2$ in segmentation tasks~\cite{miles2023mobilevos}, to align with the established practices in this field, we adopt $\alpha=2$. 
(ii) For theoretical reasons, if the application requires emphasis on tails of the distribution (rare events) or multiple modalities (distributions with multiple peaks), $\alpha$ should be less than 2 and possibly approach to 1 from above. If the goal is to highlight the dominant mode (the most probable region), $\alpha$ should be greater than 2 to emphasize central tendencies. $\alpha=2$ provides neutral weighting~\cite{yu2019multivariate}.  Moreover, the Frobenius norm's differentiable and strongly convex properties guarantee rapid convergence in gradient-based optimization algorithms~\cite{boyd2004convex}.}

 \textcolor{black}{Furthermore, in Table~\ref{tab:alpha_2}, we conduct an analysis to evaluate the performance of different $\alpha$ values ($\alpha=1.01, 2, 3$). Following with prior work~\cite{yu2019multivariate}, we set $\alpha=1.01$ to asymptotically approach Shannon entropy. The results indicate that $\alpha=2$ achieves the highest verification accuracy while reducing computational overhead by an order of magnitude. This computational gain stems from its exclusive reliance on Frobenius norm operations, whereas $\alpha=1.01$ or $3$ require eigenvalue decompositions, which are computationally more expensive.}

\begin{table}[H]
\caption{ \textcolor{black}{Experiments of different $\alpha$ values in matrix-based Rényi's entropy.}}
\label{tab:alpha_2}
\begin{center}
\begin{scriptsize}
\setlength{\tabcolsep}{4pt}
\begin{tabular}{@{}llcc@{}}
\toprule
\multirow{2}{*}{\textsc{\textbf{Method}}} & \textsc{\textbf{Agriculture}} & \textsc{\textbf{Remote Sensing}} & \textsc{\textbf{Computation Time}} \\ 
\cmidrule(l){2-4} & IoU (Leaf)  & IoU (Road)      & ms                   \\ \midrule
$\alpha=1.01$                          & 75.3\textsubscript{$\pm$ 0.31}                   & 60.6\textsubscript{$\pm$ 0.12}                & 32.1\textsubscript{$\pm$ 30.7}      \\
$\alpha=2$                         & \textbf{75.6\textsubscript{$\pm$ 0.27}}                   & \textbf{61.4\textsubscript{$\pm$ 0.30}}                & \textbf{1.2\textsubscript{$\pm$ 0.3}}      \\
$\alpha=3$                           & 75.2\textsubscript{$\pm$ 0.30}                   & 61.2\textsubscript{$\pm$ 0.06}                & 35.4\textsubscript{$\pm$ 31.2}      \\
\bottomrule
\end{tabular}
\end{scriptsize}
\end{center}
\vskip -0.2in
\end{table}

\section{Deep Dive into the Relation Model}
\label{app:deep_dive_rm}

\subsection{Understanding the Domain-invariant Information Encoded by the Relation Model}

 \textcolor{black}{Many recent studies leverage SAM's pre-trained capabilities for downstream tasks by fine-tuning. However, when the fine-tuning data distribution is narrow, the model tends to overfit task-specific local features~\cite{wang2024samcl}. We argue that this is mainly because task-specific optimizations will cover or suppress domain-invariant features learned during pre-training. }

 \textcolor{black}{To substantiate this assumption, we have conducted experiments in Section~\ref{sec:abla_study} to illustrate that the extracted relation works (see Table~\ref{tab:ablation_loss}) and is domain-invariant (see Table~\ref{tab:transfer_rm}). In Table~\ref{tab:ablation_loss}, the extracted relations boost other distillation methods (e.g., TinySAM) by 1.7\%–5.2\% IoU, indicating the preserved information's effectiveness. In Table~\ref{tab:transfer_rm}, applying the RM trained on one domain to a completely different domain still preserves its effectiveness, suggesting that these transferable relations are domain-invariant and beneficial for fine-tuning.}

\begin{figure}[H]
\vskip 0.2in
\begin{center}
\centerline{\includegraphics[width=0.75\columnwidth]{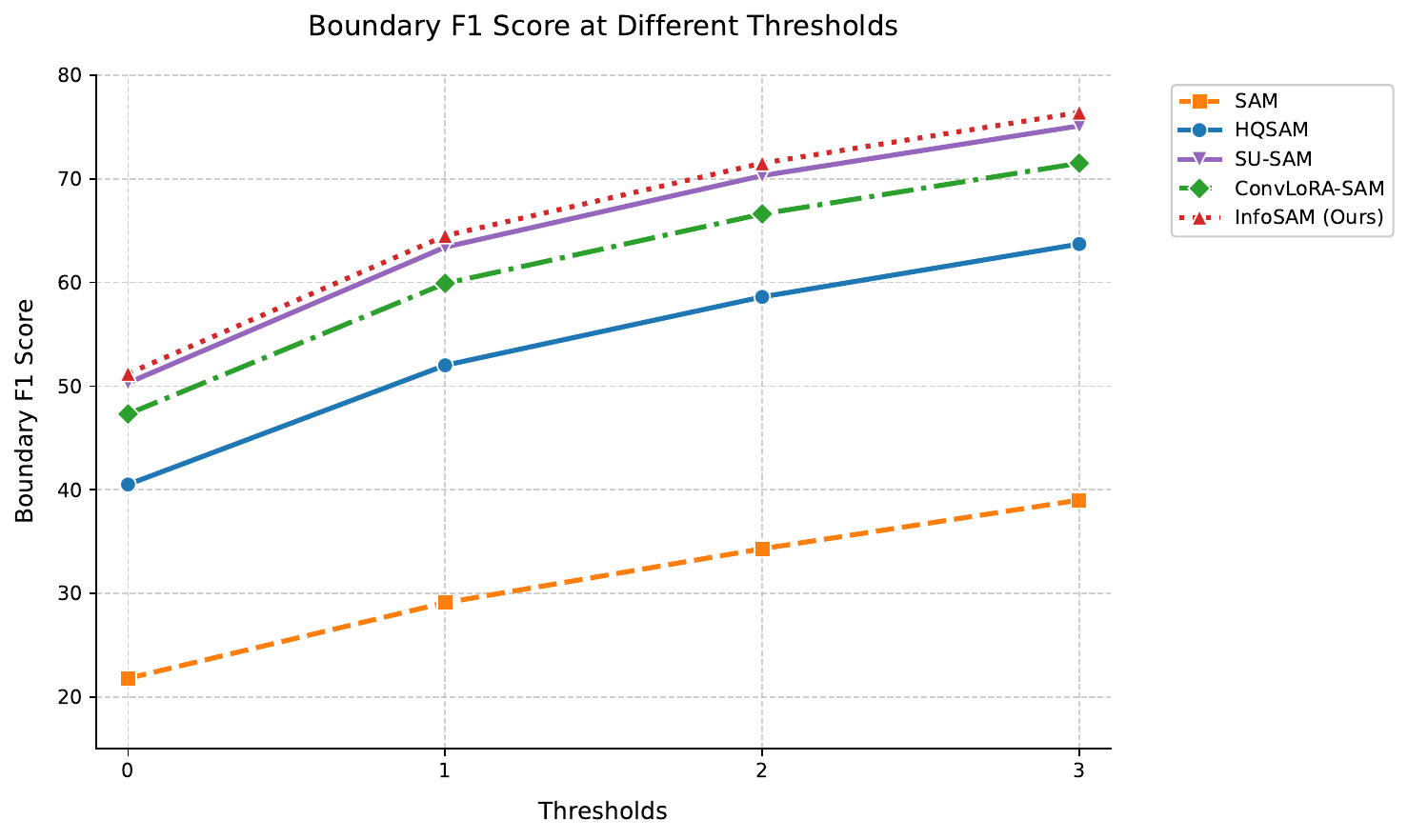}}
\caption{ \textcolor{black}{Boundary quality measured by the F1-score at various thresholds (0, 1, 2, and 3 pixels) on the Leaf dataset. The thresholds represent the allowable pixel distance: if the predicted boundary is within the threshold distance from the ground truth boundary, it is regarded as correct. A larger threshold provides a more tolerant evaluation.}}
\label{fig:boundary_f1_score}
\end{center}
\vskip -0.2in
\end{figure}

 \textcolor{black}{We further explore the nature of domain-invariant information. We employ relations to represent domain-invariant information, which serves as an implicit yet generalizable characterization that may inherently encode various domain-agnostic properties. Here, we showcase and evaluate structural edge information using the Boundary F1 Score (BFS)~\cite{zhang2023learning}. As shown in Fig.~\ref{fig:boundary_f1_score}, InfoSAM with the relation module outperforms other fine-tuning baselines in boundary preservation, demonstrating that this implicit relational encoding effectively extracts richer structural edge features.}

Additionally, we visualize the relation maps extracted from the teacher and student models at various stages of training InfoSAM, ranging from the early to late epochs.

\begin{figure}[H]
\vskip 0.2in
\begin{center}
\centerline{\includegraphics[width=0.95\columnwidth]{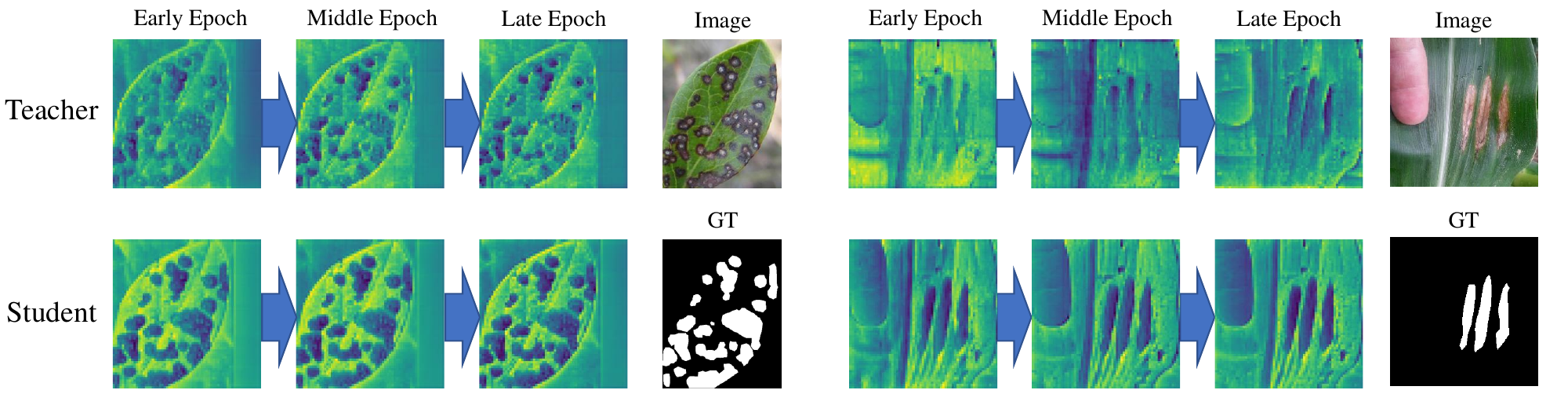}}
\caption{Relation maps evolve from early to late epochs. During training, the relation module gradually captures key information from the pre-trained teacher model, leading to improved performance of the student model.}
\end{center}
\vskip -0.2in
\end{figure}

\subsection{Effectiveness of the Proposed Loss for Relation Model Learning}

 \textcolor{black}{This section investigates the effectiveness of the proposed loss $\mathcal{L}_{\text{info}}$ in guiding the relation model to learn generalizable features while avoiding trivial solutions. As shown in Eq.~(\ref{eq_lossr}) and Eq.~(\ref{eq_loss_d}), $\mathcal{L}_{\text{info}}$ includes regularization terms such as $\log_2 \| G^{T}_{\text{imr}} \|_F^2$, $\log_2 \| G^{T}_{r} \|_F^2$, and $\log_2 \| G^{S}_{r} \|_F^2$, which promote feature diversity and prevent it from converging to trivial solutions.}

\begin{figure}[H]
\vskip 0.2in
\begin{center}
\centerline{\includegraphics[width=\columnwidth]{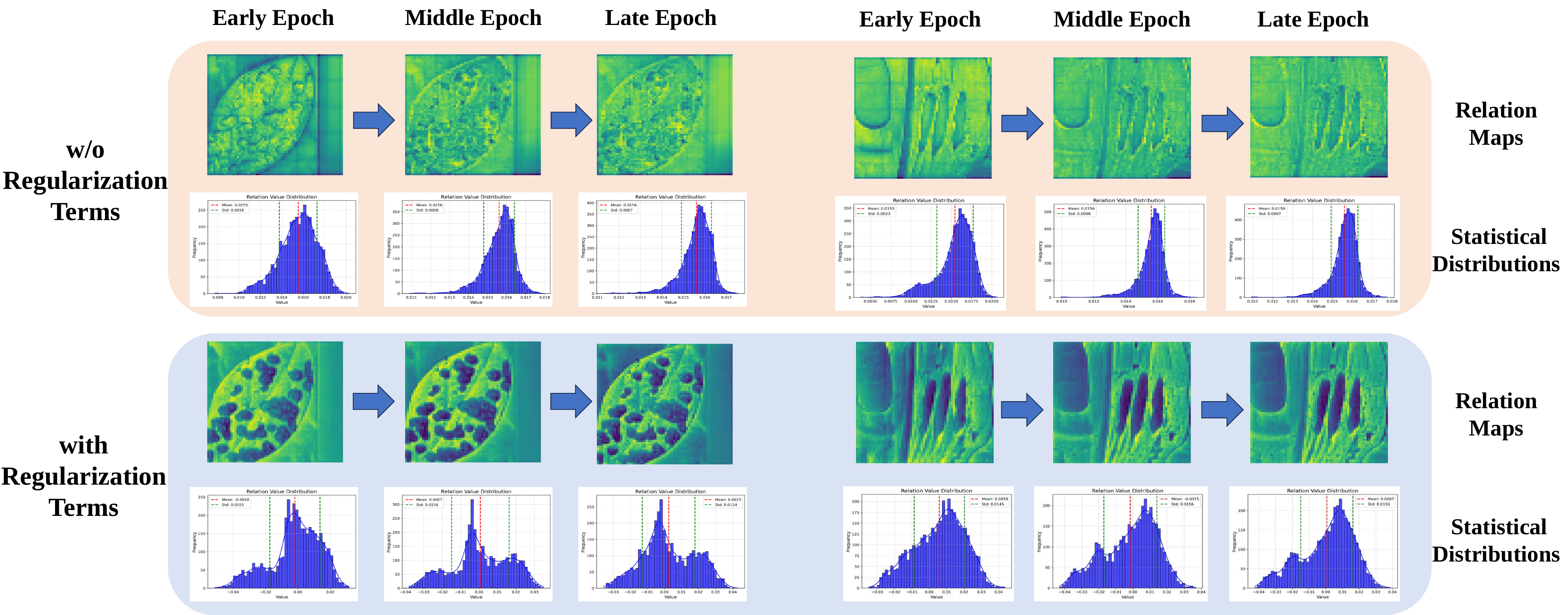}}
\caption{Evolution of relation maps and their statistical distributions over epochs, without and with the regularization term.}
\label{fig:rt_abla}
\end{center}
\vskip -0.2in
\end{figure}

 \textcolor{black}{To further verify the effectiveness of these regularization terms, we conduct an ablation study to assess their impact both qualitatively (through the visualization of relation maps) and quantitatively (through performance on downstream tasks) in Fig.~\ref{fig:rt_abla} and Table.~\ref{tab:rt_abla}, respectively: (i) For visualization performance, we visualize the relation maps and their corresponding statistical distributions evolving from early to late epochs. As shown in Fig.~\ref{fig:rt_abla}, without the regularization terms, the distribution of the relation maps becomes increasingly narrow during training, and the domain-invariant information captured by the relation maps becomes less distinct. In contrast, the RM trained with regularization terms maintains a broad relation distribution and a more representative relation map. (ii) For downstream performance shown in Table.~\ref{tab:rt_abla}, the regularization terms benefit our method by improving performance, as demonstrated by a 1.0\% and 1.8\% increase in IoU on the Leaf and Road datasets, respectively. Both results indicate that the proposed loss with regularization terms effectively extracts domain-invariant features, rather than domain-specific noise, thereby enhancing downstream performance and alleviating the problem of trivial solutions.}

\begin{table}[t]
\caption{ \textcolor{black}{Ablation study of regularization terms (RT) in $\mathcal{L}_{\text{info}}$.}}
\label{tab:rt_abla}
\begin{center}
\begin{scriptsize}
\setlength{\tabcolsep}{4pt}
\begin{tabular}{@{}llcc@{}}
\toprule
\multirow{2}{*}{\textsc{\textbf{Method}}} & \textsc{\textbf{Agriculture}} & \textsc{\textbf{Remote Sensing}} \\ 
\cmidrule(l){2-3} & IoU (Leaf)  & IoU (Road)                    \\ \midrule
w/o RT                          & 74.6\textsubscript{$\pm$ 0.12}                   & 59.6\textsubscript{$\pm$ 0.69}                  \\
w RT                         & \textbf{75.6\textsubscript{$\pm$ 0.27}}                   & \textbf{61.4\textsubscript{$\pm$ 0.30}}                    \\
\bottomrule
\end{tabular}
\end{scriptsize}
\end{center}
\end{table}

\subsection{Exploring the Relation Model architectures}
 \textcolor{black}{We conduct an analysis to compare different model architectures and explore the number of attention layers for relation module (RM). We compare direct dot product, a linear layer, multiple attention layers, and our proposed RM across multiple experiments on two distinct domains.}

\begin{table}[t]
\caption{ \textcolor{black}{Experiments of different Relation Model architectures. "Attn-$n$" represents the number of attention layers for RM.}}
\label{tab:rm_architectures}
\begin{center}
\begin{scriptsize}
\setlength{\tabcolsep}{4pt}
\begin{tabular}{@{}llcc@{}}
\toprule
\multirow{2}{*}{\textsc{\textbf{Method}}} & \textsc{\textbf{Agriculture}} & \textsc{\textbf{Remote Sensing}} \\ 
\cmidrule(l){2-3} & IoU (Leaf)  & IoU (Road)                    \\ \midrule
Dot Product                          & 75.2\textsubscript{$\pm$ 0.35}                   & 61.0\textsubscript{$\pm$ 0.04}                  \\
Linear                               & 74.9\textsubscript{$\pm$ 0.51}                   & 59.3\textsubscript{$\pm$ 0.58}                  \\
Attn-5                               & 75.4\textsubscript{$\pm$ 0.22}                   & 61.4\textsubscript{$\pm$ 0.12}                  \\
Attn-3                               & 75.4\textsubscript{$\pm$ 0.40}                   & \textbf{61.7\textsubscript{$\pm$ 0.06}}                  \\
Attn-1 (ours)                        & \textbf{75.6\textsubscript{$\pm$ 0.27}}          & 61.4\textsubscript{$\pm$ 0.30}                    \\
\bottomrule
\end{tabular}
\end{scriptsize}
\end{center}
\end{table}

 \textcolor{black}{The experimental results show that: (i) attention-based RM outperforms other other architectures designs. This indicates that attention mechanism effectively assess the correlations between the input features (i.e., image and mask features), thereby adaptively filtering and enhancing the useful information (e.g., edge details) while reducing redundancy. (ii) If we stack an appropriate number of attention layers (e.g., 3 layers) in the RM can be beneficial for capturing key information. However, stacking too many (e.g., five layers) increases training difficulty and risks overfitting. In a nutshell, the current RM design is a trade-off between performance and computational overhead, and it effectively captures the relationships between image and mask features.}

\section{Visualization Results}
\label{app_vis_results}
We present visualization results of mask predictions across various datasets using different PEFT methods for SAM (SAM, HQSAM, SU-SAM, ConvLoRA-SAM, and InfoSAM). These results further demonstrate the superiority of our proposed InfoSAM.

\begin{figure}[H]
    \centering
    \includegraphics[width=\columnwidth]{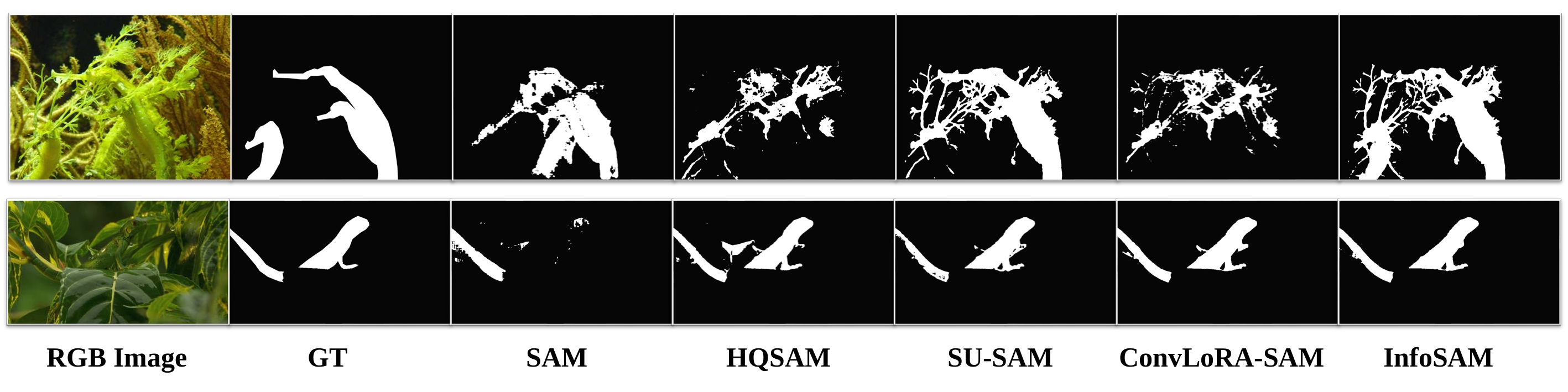}
    \caption{Visualization results on camouflaged object segmentation.}
    \label{app_fig:camo}
\end{figure}

\begin{figure}[H]
    \centering
    \includegraphics[width=\columnwidth]{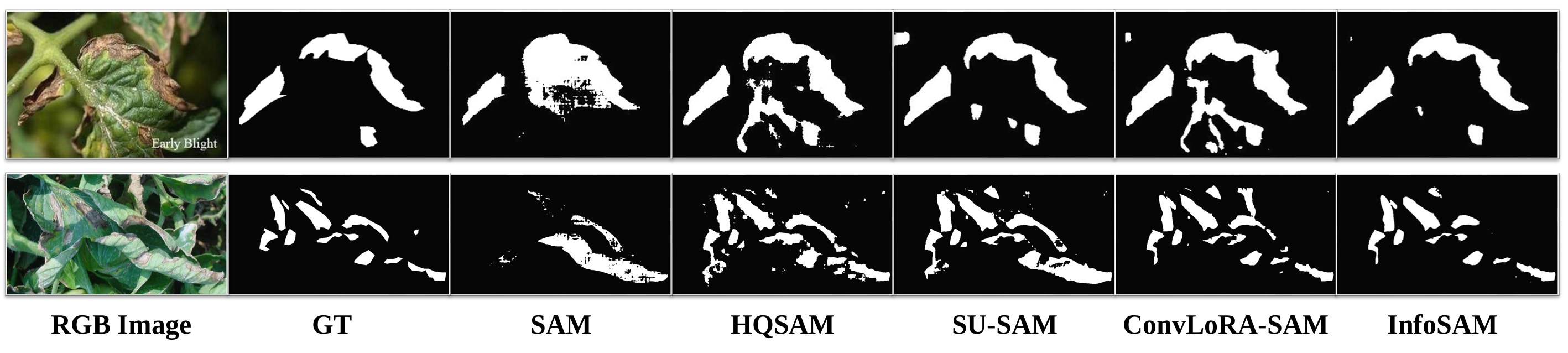}
    \caption{Visualization results on leaf disease segmentation.}
    \label{app_fig:leaf}
\end{figure}

\begin{figure}[H]
    \centering
    \includegraphics[width=\columnwidth]{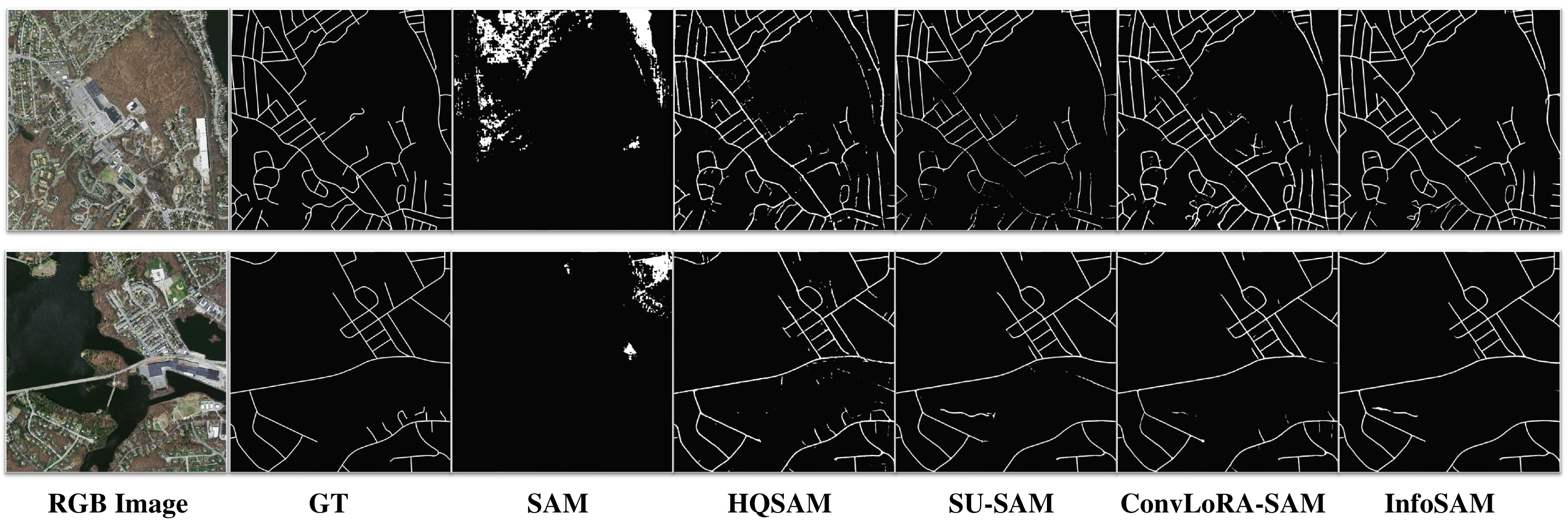}
    \caption{Visualization results on remote sensing road segmentation.}
    \label{app_fig:road}
\end{figure}

\end{document}